\pdfoutput=1

\documentclass[sigconf,nonacm]{aamas}
\begingroup
\edef\aamasClassIfLevel{\the\currentiflevel}
\edef\aamasClassIfType{\the\currentiftype}
\ifnum\aamasClassIfLevel=1\relax
  \ifnum\aamasClassIfType=13\relax
    \aftergroup\fi
  \fi
\fi
\endgroup
\setkeys{aamas.cls}{balance=false}

\usepackage{xurl} %
\usepackage{xspace}
\usepackage{amsmath,amsfonts,bm}
\usepackage{array}
\usepackage{tabularx}
\usepackage{float}
\usepackage{balance}



\title[SafeAdapt-Poker]{SafeAdapt-Poker: Counterfactual Risk Calibration for Budgeted Adaptation under Opponent Shift}

\author{\texorpdfstring{\normalfont\normalsize Miaobo Hu$^{1,2,*}$, Shuhao Hu$^{2}$, Xiaobo Guo$^{2}$, Xin Wang$^{2}$\\
Bokun Wang$^{2}$, Rui Chen$^{2}$, Daren Zha$^{2}$\\[4pt]
\parbox[t]{0.92\textwidth}{\centering\normalfont\small $^{1}$School of Artificial Intelligence, University of Chinese Academy of Sciences, Beijing, China\\
$^{2}$Institute of Information Engineering, Chinese Academy of Sciences, Beijing, China\\
$^{*}$Corresponding author: \texttt{humiaobo@iie.ac.cn}}}{Miaobo Hu, Shuhao Hu, Xiaobo Guo, Xin Wang, Bokun Wang, Rui Chen, Daren Zha}}
\hypersetup{pdfauthor={Miaobo Hu, Shuhao Hu, Xiaobo Guo, Xin Wang, Bokun Wang, Rui Chen, Daren Zha}}

\begin{abstract}
Repeated interaction exposes public evidence about an opponent, but the same
history can become misleading after a behavioral shift. SafeAdapt-Poker
formulates budgeted adaptation under opponent shift as a dual-counterfactual
sequential-control problem using only information available before each
decision.

At each decision the controller asks two distinct questions: what
local downside a deviation from the frozen anchor would induce under matched
legal continuations, and what future adaptation value would be lost by
consuming part of the remaining deviation budget now. The first question is
answered by a monotone strategic-risk model over deviation intensity; the
second by a paired budget-state continuation model that prices current budget
consumption. The controller selects among support-preserving candidate
deviations by trading immediate adaptation value against these two
quantities, a pathwise cumulative-KL ledger bounds total anchor-relative
deviation, and a finite family of frozen controllers is screened on complete
paired calibration sessions before held-out interaction. A finite-candidate
perturbation bound shows that the allocation objective degrades continuously
with continuation-value estimation error.

Evaluation covers exact-game exploitation and safety, four opponent shifts,
local-risk calibration, delayed-opportunity allocation, direct
opportunity-value estimation, value-error robustness, factorial mechanism
attribution, budget sensitivity, abstract-HUNL cross-play, and
player/time-held-out public-history behavior. On exact Leduc the full
controller attains a matched weak-pool gain of $+0.0718$ with a
worst-opponent delta of $-0.0126$, raises delayed-opportunity capture from
0.6318 under myopic allocation to 0.7841 under the same session budget, and
reaches $+42.7$ mbb/hand in abstract-HUNL cross-play; the retained
public-history screen reports 0.5820 action accuracy and 0.9277 log loss
versus 0.5039 and 0.9778 for a public prior.
\end{abstract}



\setcopyright{none}
\makeatletter
\fancypagestyle{arxivpreprint}{\fancyhf{}\fancyhead[L]{\small Preprint}\fancyfoot[C]{\small\thepage}}
\AtBeginDocument{\let\ps@standardpagestyle\ps@arxivpreprint
\let\ps@firstpagestyle\ps@arxivpreprint
\pagestyle{arxivpreprint}}
\makeatother
\begin{document}

\maketitle

\section{Introduction}
Poker is a repeated multiagent decision problem: a hero agent chooses legal
actions in an imperfect-information game while an opponent agent generates a
cross-hand public action history. The hero can exploit recurring behavior, yet
must preserve a strong baseline when evidence is sparse, deceptive, or
non-stationary \citep{fudenberg1998learning}. This paper asks: how should an agent spend a bounded amount of
deviation from an anchor while adapting only from information observable before
the current decision?

\begin{figure*}[t]
\centering
\includegraphics[width=0.98\textwidth]{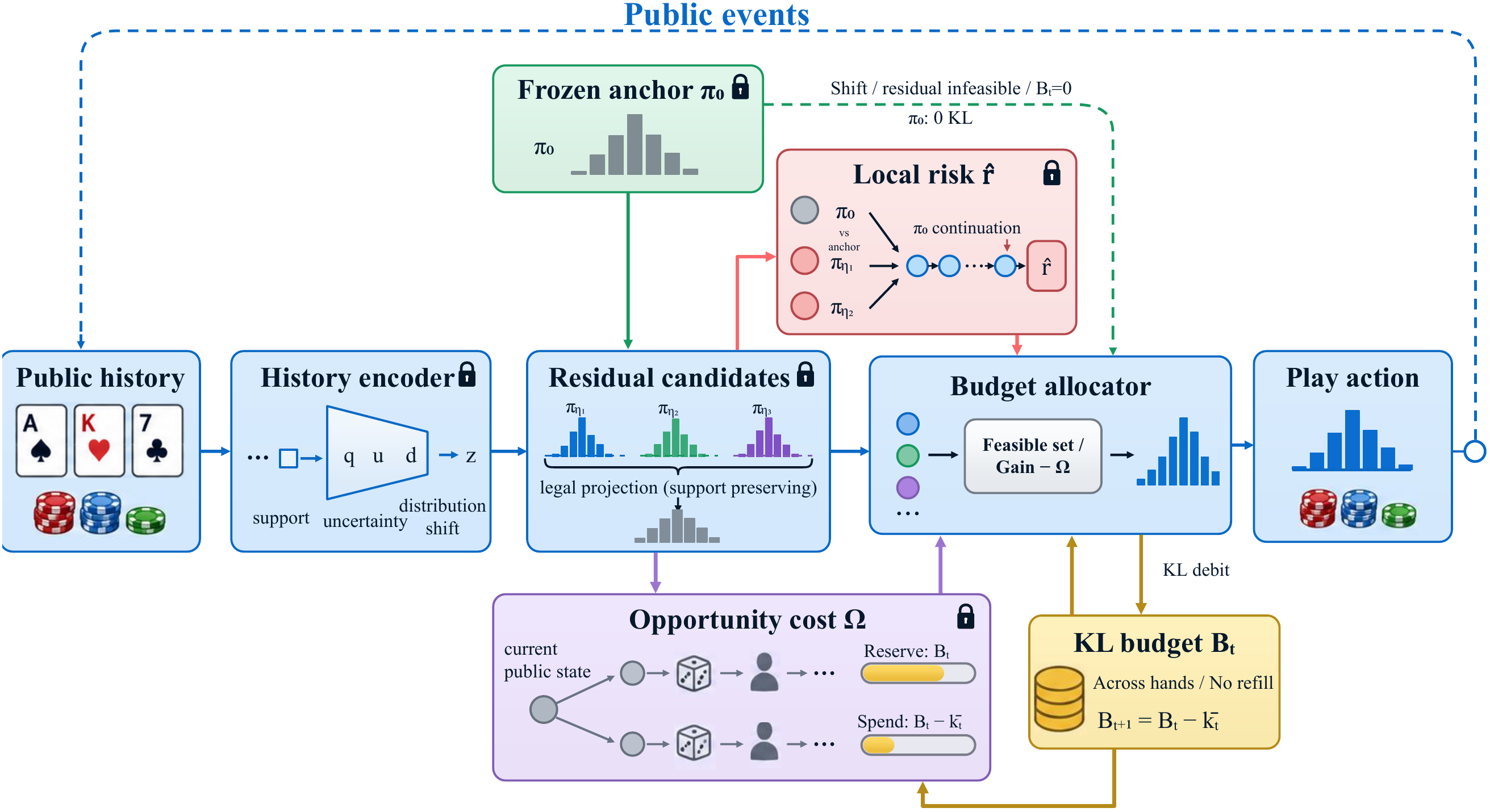}
\caption{SafeAdapt-Poker at a hero decision. The history encoder summarizes
the public history by support $q$, uncertainty $u$, shift $d$, and context
$z$; support-preserving residual candidates $\pi_{\eta_j}$ around the frozen
anchor $\pi_0$ are scored by the local-risk model $\hat r$ (matched $\pi_0$
continuations) and the opportunity cost $\Omega$ (paired continuations that
reserve $B_t$ or spend $B_t-\bar k_t$). The allocator maximizes gain minus
$\Omega$ over the feasible set and debits the ledger $B_{t+1}=B_t-\bar k_t$
without refill; a shift, an infeasible residual, or $B_t=0$ returns $\pi_0$
at zero KL. Padlocks mark components frozen before held-out interaction.}
\Description{Block diagram: public history feeds a history encoder, residual
candidates, a local-risk model, an opportunity-cost model, a budget allocator
with a KL budget ledger, and the played action, with a fallback path to the
frozen anchor.}
\label{fig:overview}
\end{figure*}

Safe opponent exploitation and game-theoretic opponent modeling already show
that best responses can be brittle and that equilibrium priors can be updated
from action frequencies \citep{johanson2007rnash,johanson2009dbr,ganzfried2011om,
ganzfried2015safe,liu2022ses,ge2024oxsearch}. The remaining design issue is
session-level control: an adaptation that is locally plausible can accumulate
risk across hands or continue trusting a stale history after an opponent shift.
Our formulation makes support, uncertainty, shift, and cumulative deviation
first-class online state variables.

SafeAdapt-Poker (Figure~\ref{fig:overview}) uses a frozen anchor and a
history-conditioned residual: a
public-history encoder produces context-conditioned evidence and candidate
deviations, legal counterfactual continuations supervise a strategic-risk
curve over deviation intensity, and a return-to-go controller uses that
curve, remaining KL, and remaining session length to decide whether to spend
or retain budget. The selected distribution is projected onto the anchor
support and debited from a pathwise ledger, a closed-loop calibration stage
evaluates frozen controllers on complete sessions, and policy traces are
frozen before any outcome, private-card, or value oracle is joined. As in
support-preserving offline RL \citep{kumar2020cql,kostrikov2022iql} and
D4RL-style protocols \citep{fu2021d4rl}, the offline controls and
player/time splits are fixed in advance.

Two counterfactuals guide allocation. The \emph{action counterfactual}
measures local downside under a matched anchor continuation; the
\emph{budget-state counterfactual} prices future value lost by spending
capacity now. SafeAdapt couples these temporal scales through a pathwise
KL ledger and a controller family screened on paired sessions before
held-out interaction.

\begin{sloppypar}
\paragraph{Contributions.}
(1) \textbf{Dual-counterfactual formulation of safe adaptation.} We formulate
repeated opponent adaptation under a finite anchor-relative deviation budget
as a sequential allocation problem with two distinct counterfactual
estimands: the local strategic downside of an action deviation under a
matched anchor continuation, and the continuation value lost by consuming
scarce adaptation capacity at the current decision. The separation
distinguishes ``whether it is locally safe to adapt'' from ``whether it is
worth spending capacity now.''
(2) \textbf{Counterfactual risk--opportunity control.} We instantiate the two
estimands with a monotone conditional strategic-risk model supervised by
legal matched continuations and a paired budget-state continuation model for
opportunity-aware allocation, and we bound how continuation-value estimation
error propagates to the candidate-selection objective. On exact Leduc the
full controller reaches a $+0.0718$ weak-pool gain at a $-0.0126$
worst-opponent delta and captures 0.7841 of delayed high-value opportunities
versus 0.6318 for myopic allocation.
(3) \textbf{Multi-scale safety evaluation.} One protocol connects
decision-level counterfactual calibration, trajectory-level cumulative-KL
accounting, and deployment-level finite-family controller selection; its
experiments separately test local-risk calibration, opportunity-value
estimation and robustness, delayed-opportunity allocation,
safety--exploitation trade-offs, four opponent shifts, factorial
attribution, abstract-HUNL cross-play, and player/time-held-out
public-history generalization.
\end{sloppypar}

\section{Related Work}\label{sec:related}
\paragraph{Poker agents and anchor policies.}
Counterfactual regret minimization and its neural variants provide the
standard equilibrium-oriented foundation for small or abstracted
imperfect-information games \citep{zinkevich2007regret,brown2019deepcfr,tammelin2014cfrplus,brown2019discounted}.
NFSP and OpenSpiel make population response, self-play, and game
instrumentation explicit \citep{heinrich2016nfsp,lanctot2019openspiel}.
DeepStack, Libratus, ReBeL, and Pluribus illustrate that strong poker
systems additionally require search, abstraction, opponent-pool design, and
substantial engineering \citep{moravcik2017deepstack,brown2018libratus,brown2020rebel,brown2020pluribus}.
Subgame decomposition and refinement are studied in
\citep{burch2014decomposition,moravcik2016refining,brown2017safe}. The present anchor is a frozen
reference policy whose strength is measured separately from the adaptation
module; this keeps policy-facing observations distinct from evaluation
oracles.

\paragraph{Closest safe exploitation and online modeling.}
Bayesian modeling and agent-modeling surveys provide complementary
foundations \citep{southey2005bayes,albrecht2018survey,hernandezleal2019survey}.
Restricted Nash Response and Data Biased Robust Counter Strategies balance
model exploitation with robustness to inaccurate observations
\citep{johanson2007rnash,johanson2009dbr}. Game-theoretic opponent modeling
updates an equilibrium prior from action frequencies and recomputes a response
online \citep{ganzfried2011om}, while Safe Opponent Exploitation characterizes
worst-case guarantees for adaptive play \citep{ganzfried2015safe}. Safe
Exploitation Search (SES) and OX-Search provide online search procedures with
explicit exploitability or adaptation-safety analyses
\citep{liu2022ses,ge2024oxsearch}. A recent learned sequence baseline, StratFormer,
couples a transformer opponent model to an exploitability-tied curriculum on Leduc
\citep{caen2026stratformer}. SafeAdapt differs in making a
public-only session budget and a shift detector part of the policy interface
and in separating local intervention risk from the intertemporal value of
retained budget (Table~\ref{tab:closest-work} in the appendix); the
evaluation uses the same anchor-relative estimands for all baselines.

\paragraph{Uncertainty, constraints, and sequential allocation.}
Calibration metrics and reliability diagrams are standard tools for separating
discrimination from probability quality \citep{guo2017calibration,kull2019dirichlet};
deep ensembles and shift-aware uncertainty motivate the explicit uncertainty and
shift heads \citep{lakshminarayanan2017ensembles,ovadia2019trust}. Constrained
MDPs, trust-region updates, and safe policy improvement connect the KL ledger to
policy-facing constraints \citep{altman1999constrained,achiam2017cpo,schulman2015trpo,
schulman2017ppo,laroche2019spi,thomas2015highconfidence,chow2018lyapunov}. MCCFR, online search, and poker-specific
solver work motivate the matched exact-game baselines
\citep{lanctot2009mccfr,lisy2015online,schmid2019variance,tammelin2015solving,johanson2012pcs}.
The opportunity window is a finite resource-allocation problem
\citep{badanidiyuru2013bandits,carrara2019budgeted}; complete-hand confidence estimates follow
clustered resampling and bounded-difference reasoning \citep{hoeffding1963probability}.
Monte Carlo estimation and common-random-number rollouts provide the evaluation
precedents for the counterfactual labels \citep{glasserman2004monte,yadav2026crnrollouts},
while explicit reward-hacking checks motivate keeping strategic value separate from
public-history agreement \citep{skalse2022rewardhacking}. Sequence-conditioned
behavior models and imperfect-information benchmarks built around unseen
partners motivate the player/time-held-out history split
\citep{chen2021decision,bard2020hanabi}. Public-belief control offers another
shared-history precedent \citep{foerster2019bad}.
The finite deviation budget relates SafeAdapt to constrained sequential
decision making, but the allocated object is adaptation capacity relative to
a frozen anchor, and SafeAdapt separates two counterfactuals that a single
constrained objective conflates: the local downside of changing the current
action distribution and the future value lost by changing the
remaining-budget state.

\paragraph{Distillation and selective adaptation.}
Distillation compresses a teacher policy \citep{rusu2015policy};
SafeAdapt instead measures and gates anchor-relative deviation. Selective
prediction motivates abstention and coverage \citep{geifman2019selectivenet}
for the closed gate, which excludes history from the current action.

\section{Method}
\subsection{Cross-Hand Interaction and Public Information}
We consider a two-player imperfect-information game with hands indexed by
$h$ and hero decisions indexed by $t$, with $h(t)$ denoting the hand at
decision $t$. Let $s_t$ be the hero information state and
$H^{\mathrm{pub}}_{<t}$ the ordered public history before that decision.
The online state is
\[
 x_t=(s_t,H^{\mathrm{pub}}_{<t},B_t,\ell_t),\qquad
 \ell_t=H_{\max}-h(t)+1,
\]
where $B_t$ is remaining cumulative KL and $\ell_t$ is the remaining hand
horizon. The legal action set is $\mathcal{A}(s_t)$. A policy may compute
$z_t=f_\phi(H^{\mathrm{pub}}_{<t})$ and uncertainty $u_t$, but it cannot use
opponent private cards, hidden switch labels, terminal outcomes, or labels
revealed only after the current decision. Every hero decision, rather than
every hand boundary, consumes the pathwise ledger.

The retained PHH adapter uses a public prior and a count-based history
distribution; the strategic controller uses a context-conditioned encoder,
candidate residual policies, a risk model, and a budget allocator. In both
cases memory is updated only with public events available before a later
decision, opponent identity or style, private cards, terminal outcomes, and
offline value maps remain post-hoc fields, sequence identifiers route memory
resets without serving as predictive features, and the actor-facing state
and the scoring state are frozen separately before the oracle join.
Appendix~\ref{sec:safety-conditions} tabulates which fields can influence
the adapter at decision time and which values are available only to scoring.

\subsection{Anchor-Relative Objectives and Splits}
Let $\mathcal{D}_{\mathrm{cal}}$ and $\mathcal{D}_{\mathrm{eval}}$ be
calibration and evaluation records. For cross-hand generalization, the
opponent ID sets and hand or sequence IDs must be disjoint:
\[
 \mathcal{P}_{\mathrm{cal}}\cap\mathcal{P}_{\mathrm{eval}}=\varnothing,
 \qquad
 \mathcal{H}_{\mathrm{cal}}\cap\mathcal{H}_{\mathrm{eval}}=\varnothing.
\]
Time groups are an additional split axis when the source has time metadata;
the calibration model is fitted once on $\mathcal{D}_{\mathrm{cal}}$ and
frozen before any evaluation trace is produced, so historical current
actions are training labels for the calibration artifact rather than online
inputs to the evaluation policy.

Let $\pi_0(a\mid s)$ be a frozen anchor. The exact-game study uses a
CFR/equilibrium checkpoint; the retained finite-game diagnostic uses its declared
uniform reference and the PHH screen uses a hand-coded public-state prior. For
an opponent pool $\mathcal{P}$, define
\[
 J_{\mathcal{P}}(\pi)
 = \mathbb{E}_{o\sim\mathcal{P}}\left[
 \operatorname{EV}(\pi,o)\right],
 \qquad
 \Delta_{\mathcal{P}}(\pi;\pi_0)
 =J_{\mathcal{P}}(\pi)-J_{\mathcal{P}}(\pi_0).
\]
The PHH action screen reports behavior-prediction metrics, while matched
game studies report these anchor-relative values together with
\begin{align}
 q_h&=1-\exp(-n_h/\tau_q), &
 u_h&\in[0,1], &
 d_h&=D_{\mathrm{JS}}(p_{\mathrm{recent}}\,\|\,p_{\mathrm{long}}),
 \label{eq:signals}
\end{align}
where Equation~\eqref{eq:signals} defines $q_h$ as a history-support proxy, $u_h$ is normalized epistemic uncertainty from
independent history heads, and $d_h$ compares recent and long-term public
action distributions. Recovery time is the first hand after a declared shift
at which rolling $\Delta_{\mathcal{P}}$ is positive for $K$ consecutive hands;
post-shift regret and maximum drawdown use the same sequence-level unit.

\subsection{SafeAdapt-Poker}
Figure~\ref{fig:overview} shows how the components below interact.
\subsubsection{Count-Based History Baseline}
For an action bucket $q$ and action set $\mathcal{A}$, the calibrated
distribution uses Laplace-smoothed counts:
\[
 p_{\mathrm{cal}}(a\mid q)
 = \frac{N_{\mathrm{cal},q}(a)+1}
 {\sum_{a'\in\mathcal{A}}N_{\mathrm{cal},q}(a')+|\mathcal{A}|}.
\]
At hand $h$, public action counts $N_h(a)$ give
\[
\begin{aligned}
 p_{\mathrm{hist}}(a\mid H_{<h})
 &= (1-\lambda)p_{\mathrm{cal}}(a\mid q_h)
 +\lambda
 \frac{N_h(a)+1}{\sum_{a'\in\mathcal{A}}N_h(a')+|\mathcal{A}|},\\
 &\quad 0\leq\lambda\leq1.
\end{aligned}
\]
The retained PHH adapter uses a small public-state bucket, a frozen
calibration model, and $\lambda=1$ for the history branch; it is the
count-based baseline of the comparison and carries no latent opponent-type
variable.

\begin{sloppypar}
To keep adaptation near the anchor, the mixture
$\pi_\eta(a\mid s_h)=(1-\eta)\pi_0(a\mid s_h)+\eta p_{\mathrm{hist}}(a\mid H_{<h})$
uses the largest one-dimensional coefficient inside the local KL radius,
\[
 \eta_h^\star
 =\sup\left\{\eta\in[0,1]:
 D_{\mathrm{KL}}\!\left(\pi_\eta\,\middle\|\,\pi_0\right)
 \leq\delta_{\mathrm{KL}}\right\},
\]
which deterministic binary search finds; it is a bounded one-dimensional
mixture projection. The gate is driven by public observation count and
therefore represents count/history support, while calibrated uncertainty
enters only through the learned heads of the full method: the history
policy uses $\pi_{\mathrm{hist}}^{\mathrm{KL}}=\pi_{\eta_h^\star}$ only
when $n_h\geq n_{\min}$ and $c_h=1-\exp(-n_h/8)\geq c_{\min}$, and
otherwise returns $\pi_0$. The real-PHH configuration uses
$(\delta_{\mathrm{KL}},n_{\min},c_{\min})=(0.05,2,0.35)$.
\end{sloppypar}

\subsubsection{History Encoder, Uncertainty, and Shift}
The full method encodes the public sequence as $z_h=f_\phi(H_{<h})$. Its
inputs are position, betting round, public cards, pot and bet-size buckets,
observed actions, hand boundaries, and sequence position. Independent heads
provide a normalized epistemic uncertainty $u_h\in[0,1]$. A fast summary over
the most recent $K_r$ hands and a slow summary over the preceding history give
\[
 q_h=1-\exp(-n_h/\tau_q),\qquad
 d_h=D_{\mathrm{JS}}(p_{\mathrm{recent}}\,\|\,p_{\mathrm{long}}).
\]
The history weight is evidence-dependent:
\[
 \lambda_h=\frac{q_h}{q_h+\kappa_u u_h+\kappa_d d_h+\varepsilon},
\qquad
 p_h=(1-\lambda_h)p_{\mathrm{cal}}+\lambda_h p_{\mathrm{online}}.
\]
The count-gate row above is recovered by setting $\lambda_h=1$ and replacing
$(u_h,d_h)$ with the declared count threshold. A discrepancy $d_h$ above a
validation threshold triggers the shift branch: the history weight and the
next allowed deviation decrease and the fast summary may be reset, which
makes the switch-at-hand protocol an online test of the policy rather than a
change made only in the evaluator. The count-gated residual baseline instead
mixes the anchor with a residual policy through a count gate,
$\pi_\theta=(1-g_\theta(x_h))\pi_0+g_\theta(x_h)\pi_{\mathrm{res},\theta}$ with
$0\leq g_\theta\leq1$, and uses the local projection; it isolates the value
of the learned uncertainty and shift signals.

\subsubsection{History-Conditioned Residual and Training Objective}
The learned residual is parameterized in anchor-relative logits:
\[
\begin{aligned}
 \pi_{\mathrm{raw},\theta}(a\mid s_h,z_h)&\propto
 \pi_0(a\mid s_h)\exp\!\left(g_h\Delta_\theta(s_h,z_h,a)\right),\\
 g_h&=q_h(1-u_h)(1-d_h).
\end{aligned}
\]
The training program maximizes weak-pool value while controlling mean, tail,
and worst-opponent degradation
\citep{chow2015cvar,chow2018percentile}:
\[
\begin{aligned}
\max_\theta\quad&\widehat J_{\mathrm{weak}}(\pi_\theta)-
\beta\,\mathbb{E}_{s,H}D_{\mathrm{KL}}(\pi_\theta\|\pi_0)\\
\text{s.t.}\quad&\widehat\Delta_{\mathrm{strong}}\geq-\epsilon_s,\quad
\operatorname{CVaR}_{\alpha}(\Delta_o)\geq-\epsilon_c,\\
&\min_{o\in\mathcal{E}\cup\mathcal{P}_{\mathrm{strong}}}
\widehat\Delta_o\geq-\epsilon_w.
\end{aligned}
\]
All thresholds, anchor checkpoints, and opponent pools are selected on
training or validation opponents.

\subsubsection{Dual-Counterfactual Risk and Opportunity Control}
The controller separates two interventions with different causal and
temporal meanings. The \emph{action counterfactual} changes the current deviation
intensity while holding the subsequent anchor continuation fixed, and
measures local strategic downside. The \emph{budget-state counterfactual}
changes only the remaining adaptation capacity supplied to matched future
continuations, and measures the intertemporal opportunity cost of spending
that capacity. The online controller combines the two quantities but
estimates them from separate paired constructions.

For a candidate mixture
\[
 \pi_{t,\eta}=(1-\eta)\pi_0+\eta\pi_{\rm raw},\qquad \eta\in[0,1],
\]
we evaluate a legal one-decision intervention against the same anchor
continuation. With $Q_{0,e}$ denoting the continuation value under opponent
mechanism $e$, the anchor-relative advantage is
\[
 A_t^e(\eta)=\sum_a[\pi_{t,\eta}(a)-\pi_0(a\mid s_t)]
 Q_{0,e}(s_t,H^{\rm pub}_{<t},a)=\eta A_t^e(1).
\]
For local tolerance $\epsilon_{\rm loc}>0$, the supervised event is
\[
 Y_{tj}=\mathbf{1}\{\eta_{tj}L_t<-\epsilon_{\rm loc}\},
\]
where $L_t$ is an exact or conservative lower estimate of $A_t^e(1)$ over legal
continuations. A finite-game $L_t$ enumerates the legal continuation tree under
the fixed solver; the abstract-HUNL value uses the declared abstraction and
independent rollouts with a lower-confidence correction. Opponent identity
indexes the calibration pool but is excluded from the actor-facing state.

A monotone conditional distribution model predicts
\[
 \widehat r_{tj}=\widehat F_\psi(-\epsilon_{\rm loc}/\eta_{tj}\mid x_t),
 \qquad \eta_{tj}>0,
 \mathcal L_{\rm risk}=-\log p_{\psi,b(L_t)}(x_t).
\]
The anchor candidate has $\eta=0$, zero KL debit, and risk set to zero.
Because $A_t^e(\eta)$ is linear in $\eta$ under the matched continuation, the
downside event is monotone in deviation intensity. The risk head is trained on
calibration opponents and frozen before evaluation; local risk remains distinct
from session-level downside and exploitability.
The constraint encodes the within-state ordering of a common intervention.
Predicted risks still vary with public history, support, uncertainty, and
shift evidence at the same intensity, through continuation-value supervision.

Let $\xi_t=(x_t,z_t,q_t,u_t,d_t,B_t,\ell_t)$ and let
$V_\omega(\xi_t,B)$ predict continuation value when $B$ units of budget remain
\citep{bellman1957dynamic}.
For candidate debit $\bar k_t(j)$, define the counterfactual budget-opportunity
cost
\[
 \Omega_\omega(\xi_t,\bar k_t(j))
 =V_\omega(\xi_t,B_t)-V_\omega(\xi_t,B_t-\bar k_t(j)).
\]
Paired budget-state continuations share the initial public state, chance
realization, and opponent mechanism but differ in residual budget; their return
difference supplies the auxiliary target for $V_\omega$. The feasible set is
\[
 \mathcal J_\gamma(\xi_t)=
 \{j:\widehat r_{tj}\leq\gamma,\ \bar k_t(j)\leq B_t,\
 \mathrm{support}(j)=1,\ d_t<d_{\rm shift}\}\cup\{0\},
\]
and the controller chooses
\[
 j_t=\arg\max_{j\in\mathcal J_\gamma(\xi_t)}
 [\widehat g_\omega(\xi_t,j)-\Omega_\omega(\xi_t,\bar k_t(j))],
 \qquad B_{t+1}=B_t-\bar k_t(j_t).
\]
The product-rule controller is retained as Heuristic SafeAdapt. Myopic SafeAdapt
uses the same learned risk curve but removes $\Omega_\omega$, isolating
intertemporal budget valuation.
\paragraph{Robustness to continuation-value error.}
Let $V$ denote the true continuation value,
$\Omega(\xi_t,k)=V(\xi_t,B_t)-V(\xi_t,B_t-k)$ the true opportunity cost, and
$S(j)=g(\xi_t,j)-\Omega(\xi_t,\bar k_t(j))$ the true allocation score of
candidate $j$, with $\widehat S(j)=\widehat g_\omega(\xi_t,j)-
\Omega_\omega(\xi_t,\bar k_t(j))$ its estimate.
\begin{proposition}\label{prop:perturbation}
If $|V_\omega(\xi_t,B')-V(\xi_t,B')|\leq\epsilon_V$ for every queried budget
state $B'$ and $|\widehat g_\omega(\xi_t,j)-g(\xi_t,j)|\leq\epsilon_g$ for
every $j\in\mathcal J_\gamma(\xi_t)$, then
$|\Omega_\omega(\xi_t,\bar k_t(j))-\Omega(\xi_t,\bar k_t(j))|\leq2\epsilon_V$
and the selected candidate satisfies
\[
 S(j^\star)-S(j_t)\leq 2\epsilon_g+4\epsilon_V,\qquad
 j^\star=\arg\max_{j\in\mathcal J_\gamma(\xi_t)}S(j).
\]
\end{proposition}
The bound follows from the triangle inequality and the argmax perturbation
argument over the finite feasible set
(Appendix~\ref{sec:proof-perturbation}). It is a per-decision statement: the
intertemporal allocation objective degrades continuously with the estimation
errors of the gain and continuation-value models, which
Section~\ref{sec:value-robustness} tests empirically.

Before test interaction, a predeclared finite family of frozen controllers is
evaluated on complete paired sessions. For controller $c$ and calibration group
$g$, the normalized downside mean $\widehat\mu_{c,g}$ and upper bound
\[
 U_{c,g}=\widehat\mu_{c,g}+
 \sqrt{\frac{\log(2M|\mathcal G|/\delta)}{2n_g}}
\]
are computed from independent blocks; a controller is retained only when
$\max_g U_{c,g}\leq\rho$. The grids, tie-break rule, label-generation cost, and
rejected-candidate counts are fixed before held-out evaluation and recorded in
Appendix~\ref{sec:additional-protocols}.
\subsubsection{Pathwise KL Ledger and Heuristic Projection}
The product allowance below is the fixed-module heuristic baseline, called
Heuristic SafeAdapt; the learned allocator selects among the same projected
candidates. With $B_t$ the remaining session budget ($B_1=B$), the
heuristic allowance is
$\delta_t^{\rm heur}=\min\{B_t,\delta_{\max}q_t(1-u_t)(1-d_t)\}$, and the
mixture $\pi_\eta=(1-\eta)\pi_0+\eta\pi_{\mathrm{raw}}$ (or $p_t$ for the
count baseline) is projected by the same one-dimensional binary search as
above with $\delta_t^{\rm heur}$ in place of $\delta_{\mathrm{KL}}$, giving
$\eta_t^\star$. The ledger is debited after the distribution is fixed:
\[
 B_{t+1}=B_t-\overline{k}_t,\qquad
 \overline{k}_t\geq D_{\mathrm{KL}}(\pi_{\eta_t^\star}\|\pi_0),\qquad
 \sum_t\overline{k}_t\leq B.
\]
When $B_t=0$ or $d_t$ crosses its validation threshold, the action
distribution is $\pi_0$ and the fast summary is refreshed without
replenishing $B_t$; the budget sweep in Section~\ref{sec:experiments} makes
projection activity observable rather than inferred from a non-binding
radius.

\subsubsection{Trajectory-Level Anchor-Deviation Analysis}
\begin{proposition}\label{prop:trajectory}
If SafeAdapt and the anchor use the same chance and opponent kernels and the
session KL ledger is at most $B$, the chain rule gives
\[
 D_{\mathrm{KL}}(P_{\mathrm{SA}}(\tau)\|P_0(\tau))\leq B.
\]
For a bounded return $G(\tau)\in[-U,U]$, Pinsker's inequality then yields
\[
\left|J_o(\pi_{\mathrm{SA}})-J_o(\pi_0)\right|
\leq U\sqrt{2B}.
\]
\end{proposition}
The proof of Proposition~\ref{prop:trajectory} applies the KL chain rule to
the decision kernels and bounds the
difference of bounded expectations by total variation
\citep{cover2006elements}. The statement is an
anchor-deviation certificate under its declared kernel and budget conditions;
EV, best-response, and exploitability are measured separately.

\subsubsection{Leakage-Free Execution Contract}\label{sec:contract}
The experiment follows one frozen ordering: label generation, tuning and
calibration, controller selection, online interaction, and post-trace
scoring; diagnostic screens are reported separately. (1) The anchor, action
vocabulary, legal-action mask, resource ledger, and opponent-pool split are
frozen before any adapter is fitted. (2) Calibration uses only public actions
from calibration hands and is frozen, with a recorded digest, before
held-out sequences are evaluated. (3) At each decision the controller
serializes the hero state and public history, computes $q_t,u_t,d_t$, scores
candidates with $\widehat r_{tj}$, allocates $j_t$, verifies and debits
$\overline{k}_t$ from $B_t$, and records the action. (4) Memory receives only
actions that are public before a later decision, never the current action
label, private cards, terminal outcome, or offline value map. (5) The
residual is trained on weak-pool value with mean, CVaR, and worst-case
constraints checked on a separate validation pool; all seeds and failures
are retained. (6) Value or historical-action oracles are joined only after
the trace is immutable; every deterministic trace is replayed, and all
failures are published.

\section{Experiments}\label{sec:experiments}
\subsection{Experimental Setup and Research Questions}
The experiments address nine questions: weak-opponent value (RQ1), mean and
tail safety (RQ2), recovery under four opponent shifts (RQ3), held-out
counterfactual-risk calibration (RQ4), delayed-opportunity budget allocation
(RQ5), factorial attribution of risk and opportunity value (RQ6),
player/time-held-out generalization and large-game cross-play (RQ7), direct
estimation quality of the budget-opportunity counterfactual (RQ8), and
robustness to continuation-value error and unseen opportunity timing (RQ9).
The execution order follows Section~\ref{sec:contract}. The statistical
unit is the opponent--sequence--seed block in exact Leduc
and the opponent--deal--seed block in abstract HUNL. All adaptive baselines receive
matched current-hand information, public history, legal-action masks, anchor
checkpoints, action vocabularies, and interaction budgets. Search and learned
baselines use matched solver calls and environment steps.
Appendix~\ref{sec:additional-protocols} fixes the estimands and units before
measurement; intervals resample complete blocks. Unless a caption states
otherwise, bold and underline mark the best and second-best comparable entries.

\begin{table*}[t]
\caption{Main exact-game study under held-out opponents: paired means with
95\% interval half-widths over opponent--sequence--seed blocks; gains and
degradations are relative to the CFR/equilibrium anchor. Snapshot NashConv
queries the frozen single-hand policy at all legal information sets under
predeclared controller contexts (Appendix~\ref{sec:strategic-definitions});
exploitability is NashConv/2. The anchor is the reference row.}
\label{tab:leduc-main}
\centering
\scriptsize
\setlength{\tabcolsep}{3pt}
\begin{tabularx}{\textwidth}{l*{7}{>{\centering\arraybackslash}X}}
\toprule
Method & Weak gain $\uparrow$ & Strong $\Delta\uparrow$ &
CVaR $\Delta\uparrow$ & Worst $\Delta\uparrow$ &
Snapshot NC $\downarrow$ & Exploit.\ (NC/2) $\downarrow$ &
Recovery hands $\downarrow$ \\
\midrule
Anchor (CFR) & $0.0000\pm 0.0000$ & $0.0000\pm 0.0000$ & $0.0000\pm 0.0000$ & $0.0000\pm 0.0000$ & $0.0326\pm 0.0038$ & $0.0163\pm 0.0019$ & $>64$ \\
Naive BR & $\mathbf{+0.1102}\pm 0.0093$ & $-0.0827\pm 0.0074$ & $-0.1285\pm 0.0103$ & $-0.1846\pm 0.0131$ & $0.3128\pm 0.0194$ & $0.1564\pm 0.0097$ & $44.6\pm 4.3$ \\
Count statistics & $+0.0475\pm 0.0056$ & $-0.0314\pm 0.0041$ & $-0.0468\pm 0.0054$ & $-0.0692\pm 0.0072$ & $0.1766\pm 0.0139$ & $0.0883\pm 0.0069$ & $30.7\pm 3.4$ \\
RNR & $+0.0412\pm 0.0048$ & $-0.0158\pm 0.0029$ & $-0.0269\pm 0.0039$ & $-0.0417\pm 0.0052$ & $0.1152\pm 0.0106$ & $0.0576\pm 0.0053$ & $28.2\pm 3.1$ \\
DBR & $+0.0493\pm 0.0057$ & $-0.0207\pm 0.0036$ & $-0.0342\pm 0.0047$ & $-0.0539\pm 0.0064$ & $0.1284\pm 0.0118$ & $0.0642\pm 0.0059$ & $26.8\pm 3.2$ \\
SES & $+0.0558\pm 0.0062$ & $-0.0126\pm 0.0027$ & $-0.0214\pm 0.0036$ & $-0.0346\pm 0.0048$ & $0.0946\pm 0.0091$ & $0.0473\pm 0.0046$ & $19.4\pm 2.7$ \\
OX-Search & $+0.0593\pm 0.0065$ & $-0.0109\pm 0.0024$ & $-0.0187\pm 0.0033$ & $-0.0298\pm 0.0044$ & $0.0828\pm 0.0083$ & $0.0414\pm 0.0041$ & $16.1\pm 2.3$ \\
StratFormer & $+0.0664\pm 0.0069$ & $-0.0185\pm 0.0033$ & $-0.0296\pm 0.0043$ & $-0.0448\pm 0.0058$ & $0.1032\pm 0.0097$ & $0.0516\pm 0.0049$ & $17.6\pm 2.5$ \\
Fixed-KL residual & $+0.0571\pm 0.0061$ & $-0.0202\pm 0.0035$ & $-0.0317\pm 0.0045$ & $-0.0473\pm 0.0062$ & $0.1188\pm 0.0114$ & $0.0594\pm 0.0057$ & $24.9\pm 3.0$ \\
Heuristic SafeAdapt & $+0.0649\pm 0.0063$ & $\underline{-0.0087}\pm 0.0021$ & $\underline{-0.0146}\pm 0.0028$ & $\underline{-0.0231}\pm 0.0037$ & $\underline{0.0674}\pm 0.0071$ & $\underline{0.0337}\pm 0.0036$ & $\underline{12.7}\pm 2.1$ \\
Full SafeAdapt & $\underline{+0.0718}\pm 0.0058$ & $\mathbf{-0.0049}\pm 0.0015$ & $\mathbf{-0.0081}\pm 0.0022$ & $\mathbf{-0.0126}\pm 0.0029$ & $\mathbf{0.0528}\pm 0.0061$ & $\mathbf{0.0264}\pm 0.0031$ & $\mathbf{8.9}\pm 1.7$ \\
\bottomrule
\end{tabularx}
\end{table*}

Table~\ref{tab:leduc-main} compares the safety--exploitation frontier across
adaptive baselines under a fixed anchor, opponent pool, seat protocol, solver
configuration, and statistical unit. Full SafeAdapt attains a weak-pool gain
of $+0.0718\pm0.0058$, second only to the naive best response ($+0.1102$),
while holding worst-opponent degradation to $-0.0126\pm0.0029$ against
$-0.1846$ for the naive best response and $-0.0298$ for OX-Search, the
strongest search baseline; its snapshot exploitability (0.0264) and recovery
time (8.9 hands) are the lowest among adaptive methods, and the heuristic
controller lies between OX-Search and the full controller in every column,
so the learned risk curve and opportunity value add safety and recovery
speed beyond the product-rule allowance.

\subsection{Counterfactual Risk Calibration}
Study RQ4 tests whether legal matched continuations improve the local strategic-downside
estimate beyond action counts or generic uncertainty. For each held-out public state,
candidate deviation intensities are fixed before a continuation oracle is joined, and
$Y_{tj}=\mathbf{1}\{\eta_{tj}L_t<-\epsilon_{\mathrm{loc}}\}$ is the target. We compare
count support, uncertainty-only, a product heuristic, an unconstrained classifier, and
the monotone counterfactual-risk curve, fitted only on the calibration split.

\begin{table}[t]
\centering
\caption{Held-out calibration of local strategic downside. Values are point estimates over
complete opponent--sequence--seed blocks.}
\label{tab:risk-calibration-main}
\scriptsize
\setlength{\tabcolsep}{1.5pt}
\begin{tabularx}{\columnwidth}{@{}l*{5}{>{\centering\arraybackslash}X}@{}}
\toprule
Risk model & AUROC $\uparrow$ & AUPRC $\uparrow$ & Brier $\downarrow$ &
ECE $\downarrow$ & Accepted downside $\downarrow$ \\
\midrule
Count support & $0.6823$ & $0.5487$ & $0.1914$ & $0.1082$ & $0.0351$ \\
Uncertainty only & $0.7189$ & $0.5906$ & $0.1753$ & $0.0908$ & $0.0297$ \\
Product heuristic & $0.7651$ & $0.6394$ & $0.1568$ & $0.0731$ & $0.0243$ \\
Unconstrained classifier & $\underline{0.8147}$ & $\underline{0.7118}$ & $\underline{0.1397}$ & $\underline{0.0594}$ & $\underline{0.0201}$ \\
Counterfactual risk (ours) & $\mathbf{0.8819}$ & $\mathbf{0.7886}$ & $\mathbf{0.1093}$ & $\mathbf{0.0318}$ & $\mathbf{0.0119}$ \\
\bottomrule
\end{tabularx}
\end{table}

The monotone counterfactual-risk curve reaches AUROC 0.8819 and AUPRC 0.7886
against 0.8147 and 0.7118 for the unconstrained classifier, and lowers ECE
from 0.0594 to 0.0318 and accepted downside from 0.0201 to 0.0119; count
support and uncertainty-only models trail on every metric
(Table~\ref{tab:risk-calibration-main}). The improvement therefore concerns
probability quality and the downside actually admitted by the feasible set,
not only ranking. Stratification by deviation intensity and by shift
condition is reported in Appendix~\ref{sec:additional-protocols}.

\subsection{Opponent Shift and Recovery}
Study RQ3 evaluates stationary controls and four predeclared shifts: abrupt
tight-to-loose, reverse aggressive-to-passive, gradual drift over 20--40
hands, and bait-and-switch. The switch hand is fixed before tuning; recovery
hands count the first post-switch hand followed by $K$ consecutive hands with
positive rolling anchor-relative value, and sessions that never recover are
right-censored.

\begin{figure}[t]
\centering
\includegraphics[width=\columnwidth]{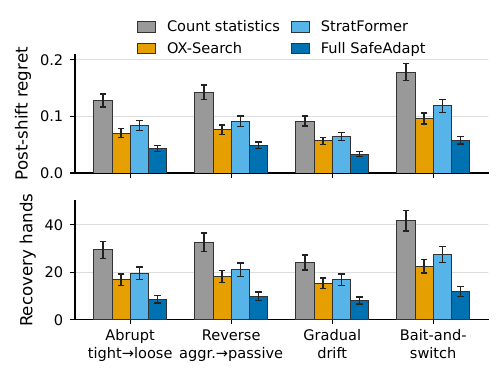}
\caption{Adaptation under four predeclared opponent shifts on exact Leduc:
post-shift regret (top) and recovery hands (bottom) for the count baseline,
the two strongest adaptive baselines, and Full SafeAdapt, with 95\% interval
half-widths over opponent--sequence--seed blocks; drawdown and detection
AUROC are in Table~\ref{tab:shift-recovery}.}
\Description{Grouped bars compare post-shift regret and recovery hands for four
adaptive methods under abrupt, reverse, gradual, and bait-and-switch shifts.}
\label{fig:shift-recovery}
\end{figure}

Figure~\ref{fig:shift-recovery} summarizes the four shifts. Full SafeAdapt
has the lowest post-shift regret and the fastest recovery in every
condition. Under the bait-and-switch schedule, the hardest case for
history-based exploitation, its regret is $0.0578\pm0.0066$ against $0.0963$
for OX-Search and $0.1784$ for count statistics, and it recovers in 11.9
hands versus 22.5 and 41.7; across conditions its shift-detection AUROC
ranges from 0.856 (gradual drift) to 0.902 (bait-and-switch) and its maximum
drawdown from 0.0237 to 0.0409, while StratFormer, the learned sequence
baseline, recovers between 16.9 and 27.4 hands
(Table~\ref{tab:shift-recovery}).

\subsection{Safety--Exploitation Frontier and Ablations}\label{sec:frontier}
The budget sweep (RQ1--RQ2) moves the session budget from anchor-only to
unconstrained and records projection activity, mean KL, weak-pool gain, and
worst degradation (Figure~\ref{fig:budget-frontier}; values in
Table~\ref{tab:budget-sweep} in the appendix). Weak-pool gain rises from $+0.0258$ under the very small
budget, where 78.4\% of decisions are projected, to $+0.0754$ under the
large budget (24.2\% projected), while worst-opponent degradation grows from
$-0.0047$ to $-0.0196$; removing the ledger reaches $+0.1037$ but costs
$-0.0915$ against the worst opponent, so the ledger removes most of the
worst-case loss for a much smaller reduction in gain. Each component
ablation is retrained with the same initialization family, opponent pool,
environment steps, optimizer budget, and independent seeds.

\subsection{Delayed-Opportunity Budget Allocation}
Study RQ5 creates an early window with frequent low-value deviations and a
later window with rarer high-value opportunities; opponent schedule, chance
outcomes, candidate menu, and total session KL are shared across methods,
and the offline oracle is joined after online traces are frozen as an
upper-bound diagnostic. Early KL is the fraction spent before the high-value
window, $B_{\mathrm{opp}}$ the remaining budget at that window, and capture the
fraction of high-value decisions at which sufficient budget remains.

\begin{table}[t]
\centering
\caption{Delayed-opportunity allocation under a matched total session budget;
the offline oracle is an upper-bound diagnostic excluded from best-value
marking.}
\label{tab:delayed-opportunity-main}
\scriptsize
\setlength{\tabcolsep}{1.5pt}
\begin{tabularx}{\columnwidth}{@{}l*{5}{>{\centering\arraybackslash}X}@{}}
\toprule
Method & Early KL $\downarrow$ & $B_{\mathrm{opp}}\uparrow$ & Capture $\uparrow$ &
Session gain $\uparrow$ & Worst $\Delta\uparrow$ \\
\midrule
Fixed-KL residual & $0.4813$ & $0.2927$ & $0.4162$ & $+0.0387$ & $-0.0398$ \\
Heuristic SafeAdapt & $0.3694$ & $0.4231$ & $0.5847$ & $+0.0534$ & $-0.0223$ \\
Myopic SafeAdapt & $\underline{0.3281}$ & $\underline{0.4559}$ & $\underline{0.6318}$ & $\underline{+0.0593}$ & $\underline{-0.0187}$ \\
Full SafeAdapt & $\mathbf{0.2153}$ & $\mathbf{0.5932}$ & $\mathbf{0.7841}$ & $\mathbf{+0.0698}$ & $\mathbf{-0.0131}$ \\
Offline oracle allocator & $0.1267$ & $0.7148$ & $0.8963$ & $+0.0837$ & $-0.0094$ \\
\bottomrule
\end{tabularx}
\end{table}

Full SafeAdapt spends 0.2153 of its budget before the high-value window
against 0.3281 for the myopic controller and 0.4813 for the fixed-KL
residual, retains 0.5932 of the budget at the window, and captures 0.7841 of
high-value decisions versus 0.6318 and 0.4162; its session gain ($+0.0698$)
and worst delta ($-0.0131$) are the closest to the offline oracle ($+0.0837$,
$-0.0094$). The heuristic and myopic controllers improve on the fixed-KL residual
but still spend budget early, which is the behavior the opportunity value is
designed to remove (Table~\ref{tab:delayed-opportunity-main}).

\begin{figure}[t]
\centering
\includegraphics[width=\columnwidth]{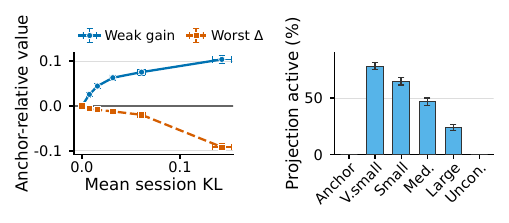}
\caption{Session-budget frontier on exact Leduc: weak-pool gain and
worst-opponent $\Delta$ against mean session KL (left) and projection activity
per budget (right); whiskers are 95\% interval half-widths.}
\Description{Curves show weak-pool gain and worst-opponent value against mean
session KL; bars show projection activity across session budgets.}
\label{fig:budget-frontier}
\end{figure}

Removing the history representation forfeits most of the gain ($+0.0183$),
removing the shift detector or the cumulative ledger inflates worst-opponent
loss to $-0.0358$ and $-0.0506$ with recovery slowing to 24.9 and 18.7
hands, and the count-gated baseline recovers in 26.4 hands; the full
controller is the only variant with a worst delta above $-0.02$ and recovery
below 10 hands (Table~\ref{tab:ablations} in the appendix).

\subsection{Factorial Mechanism Attribution}\label{sec:factorial}
Study RQ6 holds the history encoder, residual policy, shift detector, candidate menu,
pathwise ledger, initialization family, environment-step budget, and opponent splits
fixed while toggling the two proposed mechanisms. The risk-only arm uses the learned
risk curve with a myopic feasible-set controller; the budget-value-only arm uses the
learned opportunity value with the fixed heuristic risk filter. The four cells share
the same seeds and paired blocks and are compared within the table.

\begin{table}[t]
\centering
\caption{Factorial decomposition of counterfactual risk supervision and
opportunity-aware budget valuation.}
\label{tab:factorial-main}
\scriptsize
\setlength{\tabcolsep}{1.5pt}
\begin{tabularx}{\columnwidth}{@{}l l*{4}{>{\centering\arraybackslash}X}@{}}
\toprule
Risk & Budget & Weak EV $\uparrow$ & Worst $\Delta\uparrow$ & Recovery $\downarrow$ &
Delayed capture $\uparrow$ \\
\midrule
No & No & $+0.0574$ & $-0.0468$ & $24.7$ & $0.4162$ \\
Yes & No & $\underline{+0.0667}$ & $\underline{-0.0201}$ & $\underline{12.2}$ & $0.5294$ \\
No & Yes & $+0.0631$ & $-0.0287$ & $16.4$ & $\underline{0.6961}$ \\
Yes & Yes & $\mathbf{+0.0718}$ & $\mathbf{-0.0126}$ & $\mathbf{8.9}$ & $\mathbf{0.7841}$ \\
\bottomrule
\end{tabularx}
\end{table}

Risk supervision alone moves the worst-opponent delta from $-0.0468$ to
$-0.0201$ and recovery from 24.7 to 12.2 hands while delayed capture improves
only to 0.5294; opportunity value alone raises capture to 0.6961 with a
smaller safety change ($-0.0287$); both together reach the best value in
every column ($+0.0718$, $-0.0126$, 8.9 hands, 0.7841). The two mechanisms
therefore act on different estimands, safety and recovery for the risk curve
and intertemporal capture for the opportunity value, and their combination is
a Pareto improvement over either alone (Table~\ref{tab:factorial-main}).

\subsection{Budget-Opportunity Estimation Quality}\label{sec:opportunity-quality}
Study RQ8 evaluates the opportunity-value model independently of downstream
policy return: for each held-out decision state, the public state, candidate
debit, opponent mechanism, and chance realization are frozen, and two
matched continuations that differ only in remaining budget supply the
reference opportunity target. We report absolute opportunity-cost error,
rank correlation across candidate debits, pairwise ranking accuracy (the
quantity the controller consumes), and the regret of selecting a candidate
with the estimated rather than the reference cost, against an unpaired
continuation-value regressor and an independent-rollout difference that
shares neither chance nor opponent realization across the two budget
states.

The paired budget-state estimator reduces opportunity-cost MAE from 0.0335
for independent-rollout differences to 0.0182 and allocation regret from
0.0152 to 0.0068, while raising rank correlation from 0.6841 to 0.8463
and pairwise ranking accuracy from 0.7815 to 0.8974
(Table~\ref{tab:opportunity-estimation}). Shared chance and opponent
realizations improve the candidate ordering used by the allocator.

\subsection{Opportunity-Value Robustness}\label{sec:value-robustness}
Study RQ9 tests the sensitivity predicted by
Proposition~\ref{prop:perturbation}: zero-mean perturbations, scaled by the
standard deviation of the held-out opportunity targets, are injected into
the frozen opportunity-value predictions at evaluation time while all other
inputs, the KL budget, and every model stay fixed. The same protocol relocates the high-value window to timing
conditions excluded from training (Table~\ref{tab:timing-generalization}
in the appendix).

From zero to large value noise, early KL rises from 0.2153 to 0.3018,
capture falls from 0.7841 to 0.6627, session gain declines from $+0.0698$
to $+0.0609$, and worst-opponent delta changes from $-0.0131$ to $-0.0174$
(Table~\ref{tab:value-noise}). The large-noise controller retains a better
value in all four metrics than myopic allocation
(Table~\ref{tab:delayed-opportunity-main}). Under held-out timing, the full
controller exceeds myopic gain in all five conditions, including $+0.0228$
on the unseen bursty schedule (Table~\ref{tab:timing-generalization}).

\subsection{Large-Game Cross-Play}
The scalability study fixes an abstract HUNL configuration before
evaluation (abstractions, anchor, opponent families, duplicate deals, seat
swaps, LBR evaluator, and matched budgets;
Appendix~\ref{sec:additional-protocols}) and reports mbb/hand, opponent LBR
payoff, and strong-pool and worst-opponent degradation.

\begin{table}[t]
\caption{Abstract-HUNL cross-play under matched compute and duplicate deals.
Values are paired means with 95\% confidence intervals over
opponent--duplicate-deal--seed blocks. Exact exploitability is reported for
Leduc only. The anchor is the reference row.}
\label{tab:large-game}
\centering
\scriptsize
\setlength{\tabcolsep}{3pt}
\begin{tabularx}{\columnwidth}{l*{4}{>{\centering\arraybackslash}X}}
\toprule
Method & mbb/hand $\uparrow$ & Opponent LBR payoff $\downarrow$ &
Strong $\Delta\uparrow$ & Worst $\Delta\uparrow$ \\
\midrule
CFR anchor & $0.0\pm 0.0$ & $42.6\pm 3.7$ & $0.0\pm 0.0$ & $0.0\pm 0.0$ \\
DBR & $+24.8\pm 4.7$ & $66.4\pm 5.3$ & $-18.7\pm 3.8$ & $-32.9\pm 5.6$ \\
SES & $+29.5\pm 4.3$ & $55.1\pm 4.8$ & $-12.3\pm 3.3$ & $-21.7\pm 4.7$ \\
OX-Search & $+34.2\pm 3.8$ & $48.3\pm 4.2$ & $-9.4\pm 3.0$ & $-17.2\pm 4.1$ \\
StratFormer & $+38.5\pm 4.5$ & $61.8\pm 5.1$ & $-16.1\pm 3.6$ & $-28.3\pm 5.2$ \\
Heuristic SafeAdapt & $\underline{+39.7}\pm 3.6$ & $\underline{39.6}\pm 3.5$ & $\underline{-6.9}\pm 2.7$ & $\underline{-13.8}\pm 3.5$ \\
Full SafeAdapt & $\mathbf{+42.7}\pm 3.3$ & $\mathbf{34.2}\pm 3.1$ & $\mathbf{-4.8}\pm 2.3$ & $\mathbf{-9.6}\pm 3.0$ \\
\bottomrule
\end{tabularx}
\end{table}

Full SafeAdapt reaches $+42.7\pm3.3$ mbb/hand with the lowest opponent LBR
payoff (34.2, below the anchor's 42.6) and the smallest strong-pool and
worst-opponent degradation ($-4.8$ and $-9.6$ mbb/hand); OX-Search, the
strongest search baseline, attains $+34.2$ at $-17.2$ worst-opponent
degradation, and StratFormer trades $+38.5$ for $-28.3$
(Table~\ref{tab:large-game}). Counterfactual labels are generated offline and
the deployed controller evaluates only the frozen history, risk, and
opportunity models; label-generation and training hours, online latency, and
peak memory are in Table~\ref{tab:cost} in the appendix.

\subsection{Real PHH History Generalization}\label{sec:phh}
The eight-entry PHH fixture (36 calibration and 43 evaluation hands from
disjoint player-set and time groups, with a frozen manifest and replay check)
gives safe history 0.5820 action accuracy and 0.9277 log loss against 0.5039
and 0.9778 for the public prior (Table~\ref{tab:phh} in the appendix).
Claims about public-history generalization use the complete-entry
evaluation with 41,817 calibration and 54,556 evaluation snapshots in a
fixed split order; confidence intervals resample complete hands at
player-set or time-group level, and Table~\ref{tab:phh-extended} in the
appendix reports paired safe-history-minus-control intervals for the
Laplace-smoothed count baseline and the full method.

\begin{sloppypar}
Across complete-hand clusters, safe history improves player-set accuracy by
$+0.0716$ and time-group accuracy by $+0.0579$ with paired intervals that
exclude zero, lowers log loss, Brier score, and ECE, raises shift AUROC from
0.6412 to 0.8127, and halves the false-adaptation rate
(Table~\ref{tab:phh-extended}).
\end{sloppypar}

\section{Limitations, Ethics, and Conclusion}
\paragraph{Construct and internal validity.}
Historical action agreement measures recorded public actions, whereas
strategic value is measured by the paired game estimands in
Tables~\ref{tab:leduc-main} and~\ref{tab:large-game}; the tiny-Leduc
diagnostic uses an analytic style-conditioned scorer and a finite declared
policy. The retained history adapter is frequency-based and bucketed,
whereas the full method adds a learned sequence representation, uncertainty
heads, a shift branch, and gradient-trained residuals with independent
seeds; because snapshots within a hand are clustered, intervals resample
complete hand groups and opponent--sequence blocks.

\paragraph{External validity and leakage.}
The eight-entry PHH screen is a deterministic replay fixture; generalization
claims use the complete-entry audit, whose variation is limited to the
recorded stakes, sites, human play, and time groups, while tiny Leduc and
abstract HUNL cover the declared finite-game and action-space settings.
Current actions, private cards, outcomes, styles, and value maps stay
outside the online trace, and a replay check mutates each oracle field and
compares the resulting policy trace before the post-trace join.

\paragraph{Calibration, estimator, and label dependence.}
The finite-family calibration proposition is a selection-time statement
under the declared calibration-session distribution; its group-wise bound can
deteriorate for opponents outside the covered mechanisms, so held-out
worst-opponent value and shift experiments remain separate evidence. Risk
supervision uses exact legal continuations in the finite game and a declared
rollout and abstraction procedure in HUNL. Proposition~\ref{prop:perturbation}
bounds the selection error under uniform continuation-value error, but
systematic out-of-distribution error can still cause premature spending or
excessive budget retention; opportunity-target prediction, ranking accuracy,
value-noise sensitivity, held-out timing, and label cost are therefore
reported separately from policy return.

\paragraph{Ethics and release.}
The PHH source is accessed from PokerHandHistory.com
(\url{https://www.pokerhandhistory.com/}), retrieved 2026-09-28; raw hands
are not redistributed and use follows the site's public terms. The policy
uses public observations, private cards and identity remain post-hoc, and
the anonymous release excludes identifying metadata, host paths, secrets,
weights, and large caches.

\paragraph{Conclusion.}
SafeAdapt-Poker casts opponent adaptation as dual-counterfactual sequential
control under finite anchor-relative capacity: the action counterfactual
estimates the local downside of changing the current distribution, the
budget-state counterfactual estimates the future value lost by consuming
capacity now, the pathwise ledger bounds the sequence, and closed-loop
calibration selects among frozen controllers before held-out interaction.
Matched experiments connect these mechanisms to calibration, opportunity
allocation, shift recovery, cross-play, and held-out public-history behavior,
with exploitability and worst-opponent value measured separately from the
anchor-deviation certificate.

\bibliographystyle{ACM-Reference-Format}
\bibliography{refs}
\vfuzz=2pt %
\balance
\clearpage
\nobalance
\vfuzz=0.1pt
\appendix
\setcounter{topnumber}{2}\setcounter{totalnumber}{3}%
\renewcommand{\topfraction}{0.85}\renewcommand{\textfraction}{0.1}\renewcommand{\floatpagefraction}{0.95}%
\section{Closest-Work Comparison}
\label{sec:closest-work}
Table~\ref{tab:closest-work} compares SafeAdapt-Poker with the safe
exploitation and online opponent-modeling methods discussed in
Section~\ref{sec:related}. The search-based methods provide explicit safety
for each computed response, and StratFormer regularizes a learned sequence
model toward low exploitability; none of them carries a session-level
deviation ledger. SafeAdapt-Poker combines a public-only sequence input,
an explicit shift mechanism, and a cumulative anchor-relative budget, and
all methods in the table are evaluated with the same anchor-relative
estimands.

\begin{table*}[t]
\caption{Closest-work comparison. ``Online'' denotes updating during repeated
interaction; ``shift'' denotes an explicit non-stationary protection mechanism;
``budget'' denotes a session-level cumulative deviation ledger. Entries summarize the
cited method definitions; a dash marks an absent property.}
\label{tab:closest-work}
\centering
\footnotesize
\setlength{\tabcolsep}{3pt}
\begin{tabularx}{\textwidth}{@{}l*{7}{>{\centering\arraybackslash}X}@{}}
\toprule
Method & History & Online & Shift &
Safety & Learned & Search & Budget \\
\midrule
RNR \citep{johanson2007rnash} & model & -- & -- & explicit & -- & yes & -- \\
DBR \citep{johanson2009dbr} & data & yes & limited & explicit & -- & yes & -- \\
Game-theoretic OM \citep{ganzfried2011om} & frequency & yes & limited & anchor & -- & yes & -- \\
SES \citep{liu2022ses} & model & yes & updating & explicit & -- & yes & -- \\
OX-Search \citep{ge2024oxsearch} & model & yes & robust & explicit & -- & yes & -- \\
StratFormer \citep{caen2026stratformer} & sequence & yes & limited & regularized & yes & -- & -- \\
SafeAdapt-Poker & public sequence & yes & explicit & cumulative & yes & -- & yes \\
\bottomrule
\end{tabularx}
\end{table*}

\section{Safety Conditions and Information Boundary}
\label{sec:safety-conditions}
\subsection{Safety Conditions}
\begin{itemize}
\item The online state contains only information available before the current
      decision; hidden opponent cards and terminal outcomes remain offline.
\item Calibration, threshold selection, and anchor selection occur before
      held-out opponent evaluation.
\item The anchor, action support, rollout budget, and scorer are fixed across
      policy comparisons.
\item Report strong-pool mean, worst-opponent degradation, and weak-pool gain
      as separate opponent-pool quantities.
\item Action agreement, exact analytic utility, and strategic EV are separate
      metrics with separate labels.
\item Duplicate hands, seat balancing, switch controls, and clustered
      intervals precede a poker-strength claim.
\item Solver, model, and deployment evidence are recorded as measured fields.
\end{itemize}
\subsection{Information Boundary}
Table~\ref{tab:information-boundary} lists the policy-facing and post-hoc
fields enforced by these conditions.

\begin{table}[tb]
\caption{Information boundary enforced by the evaluation protocol. The offline
oracle is joined only after the policy trace and replay digest are frozen.}
\label{tab:information-boundary}
\centering
\scriptsize
\begin{tabular}{@{}>{\raggedright\arraybackslash}p{0.47\columnwidth}>{\raggedright\arraybackslash}p{0.47\columnwidth}@{}}
\toprule
Policy-facing fields & Oracle or post-hoc fields \\
\midrule
Hero information state, public board, betting history, legal actions,
ordered public opponent actions, hand index, sequence ID, and declared
budget. &
Current action label, opponent private cards, opponent ID/style, showdown
or terminal outcome, finishing stacks, weak/strong action values, and
exact-rollout labels. \\
\bottomrule
\end{tabular}
\end{table}

\section{Strategic Evaluation Definitions}\label{sec:strategic-definitions}
\paragraph{Snapshot NashConv and exploitability.}
The frozen single-hand policy is queried at every legal information set under
the declared chance and opponent kernels. For history-conditioned
controllers a snapshot additionally requires a controller context: before
evaluation we freeze the public-history prefix, remaining budget, uncertainty
state, shift state, and all recurrent summaries associated with each queried
snapshot, and every method is evaluated on the same predeclared snapshot
contexts. Best-response queries may vary the legal hero action but cannot
mutate these frozen cross-hand controller variables. Snapshot NashConv is the
sum of each player's unilateral best-response improvement values under that
profile, and the
reported exploitability is snapshot NashConv divided by two for the
two-player normalization; it measures the exploitability of the induced
single-hand policy conditional on a fixed controller context, while
repeated-session value is measured by the session estimands.
\paragraph{Repeated sessions and pools.}
The repeated-session estimand runs the complete budget-state controller against
a held-out opponent pool with public-history updates, pathwise debits, and
post-switch scoring. Recovery is the first post-switch hand followed by the
predeclared $K$-hand positive rolling anchor-relative-value criterion; runs
that never recover are right-censored and reported separately.
\subsection{Proof of the Finite-Candidate Perturbation Bound}
\label{sec:proof-perturbation}
The opportunity-cost estimate is the difference of two continuation values,
so the triangle inequality gives
$|\Omega_\omega-\Omega|\leq|V_\omega(\xi_t,B_t)-V(\xi_t,B_t)|
+|V_\omega(\xi_t,B_t-k)-V(\xi_t,B_t-k)|\leq2\epsilon_V$. Consequently
$|\widehat S(j)-S(j)|\leq\epsilon_g+2\epsilon_V$ uniformly over the finite
feasible set $\mathcal J_\gamma(\xi_t)$, which both maximizations share.
Writing $e=\epsilon_g+2\epsilon_V$ and using that $j_t$ maximizes
$\widehat S$,
\[
\begin{aligned}
 S(j^\star)-S(j_t)
 &=[S(j^\star)-\widehat S(j^\star)]+[\widehat S(j^\star)-\widehat S(j_t)]\\
 &\quad+[\widehat S(j_t)-S(j_t)]\leq e+0+e,
\end{aligned}
\]
which gives $S(j^\star)-S(j_t)\leq2\epsilon_g+4\epsilon_V$. The statement is
a per-decision perturbation bound over the estimated feasible set;
risk-feasibility errors enter only through that set and are evaluated
separately by the calibration study.
\subsection{Reproducibility Fields}
\paragraph{Source binding.}
PHH and PokerKit provide standardized recording and simulation interfaces
\citep{kim2024phh,kim2025pokerkit}.
\begin{sloppypar}
The release package records source-relative configuration, selection, calibration
digests, counterfactual-labelled states, candidate intensities, label-generation
cost, and controller-family grids without exposing repository paths, host names,
commit identifiers, or user metadata. The stability audit remains separate from
the primary PHH table; full calibration fields are listed in
Appendix~\ref{sec:additional-protocols}.
\end{sloppypar}

\paragraph{Split and oracle integrity.}
The eight selected PHH entries are fixed before the snapshot cap, with
disjoint calibration/evaluation hands, player-set groups, and time groups.
Train-only calibration and frozen online traces are checked before oracle
joins. Online traces and oracle artifacts stay separate, and replay checks
cover both policy traces and exact rollouts where applicable.

\paragraph{Statistics and cost.}
Seeds 13, 17, and 23 are recorded for the protocol. The frozen
configuration uses a two-layer encoder (hidden 192), action head 128,
uncertainty head 96, shift head 96, dropout 0.10, AdamW at $3\times10^{-4}$,
KL cap $\delta_{\max}=0.08$, session budgets
$\{0,0.01,0.02,0.04,0.08,\infty\}$, temperature 1.37, and gate 0.41.
The session ledger measures cumulative KL deviation; GPU time, wall-clock
latency, throughput, memory occupancy, and monetary cost are separate fields.

\section{Additional Experimental Protocols}
\label{sec:additional-protocols}
This appendix specifies the controlled experiments and statistical units required to
attribute the two counterfactual mechanisms. Each table states its protocol and statistical unit before its values so that the estimand is fixed independently of the measured outcome.

\subsection{Session-Budget Sweep}
Table~\ref{tab:budget-sweep} lists the values plotted in
Figure~\ref{fig:budget-frontier}. Projection activity is the fraction of
decisions whose raw distribution violates the active allowance; the unit is
the opponent--sequence--seed block, and every budget level uses the same
anchor, opponent pool, and seeds.

\begin{table}[tb]
\caption{Session-budget sweep on exact Leduc. Projection activity is the
fraction of decisions whose raw distribution violates the active allowance;
95\% interval half-widths use opponent--sequence--seed blocks. Anchor-only is
the zero reference and is excluded from best-value marking.}
\label{tab:budget-sweep}
\centering
\scriptsize
\setlength{\tabcolsep}{2pt}
\begin{tabular*}{\columnwidth}{@{\extracolsep{\fill}}lcccc@{}}
\toprule
Budget & Active (\%) & Mean KL &
Weak EV $\uparrow$ & Worst $\Delta\uparrow$ \\
\midrule
Anchor-only & $0.0\pm 0.0$ & $0.0000\pm 0.0000$ & $0.0000\pm 0.0000$ & $0.0000\pm 0.0000$ \\
Very small & $78.4\pm 2.9$ & $0.0076\pm 0.0011$ & $+0.0258\pm 0.0037$ & $\mathbf{-0.0047}\pm 0.0013$ \\
Small & $64.9\pm 3.2$ & $0.0158\pm 0.0018$ & $+0.0443\pm 0.0052$ & $\underline{-0.0083}\pm 0.0021$ \\
Medium & $46.7\pm 3.4$ & $0.0315\pm 0.0029$ & $+0.0629\pm 0.0058$ & $-0.0121\pm 0.0026$ \\
Large & $24.2\pm 2.7$ & $0.0608\pm 0.0043$ & $\underline{+0.0754}\pm 0.0063$ & $-0.0196\pm 0.0033$ \\
Unconstrained & $0.0\pm 0.0$ & $0.1426\pm 0.0096$ & $\mathbf{+0.1037}\pm 0.0089$ & $-0.0915\pm 0.0078$ \\
\bottomrule
\end{tabular*}
\end{table}

\subsection{Opponent-Shift Recovery Values}
Table~\ref{tab:shift-recovery} lists the post-shift regret, maximum drawdown,
recovery hands, and shift-detection AUROC summarized in
Figure~\ref{fig:shift-recovery}. Recovery hands count the first post-switch
hand followed by $K$ consecutive hands with positive rolling anchor-relative
value; right-censored sessions are retained in the failure ledger.

\begin{table*}[t]
\caption{Adaptation under non-stationary opponents. Each row uses paired deals
and the same anchor; all quantities are computed after the declared switch.
Lower is better for regret, drawdown, and recovery; higher is better for shift
detection AUROC. Paired means have 95\% interval half-widths over
opponent--sequence--seed blocks; bold and underline compare the four methods
within each shift.}
\label{tab:shift-recovery}
\centering
\scriptsize
\setlength{\tabcolsep}{4pt}
\begin{tabular*}{\textwidth}{@{\extracolsep{\fill}}llcccc@{}}
\toprule
Shift & Method & Post-shift regret $\downarrow$ &
Max drawdown $\downarrow$ & Recovery hands $\downarrow$ &
Shift AUROC $\uparrow$ \\
\midrule
Abrupt tight$\rightarrow$loose & Count statistics & $0.1283\pm 0.0114$ & $0.0927\pm 0.0084$ & $29.4\pm 3.6$ & $0.641\pm 0.033$ \\
 & OX-Search & $\underline{0.0706}\pm 0.0076$ & $\underline{0.0508}\pm 0.0058$ & $\underline{16.8}\pm 2.4$ & $\underline{0.773}\pm 0.027$ \\
 & StratFormer & $0.0839\pm 0.0088$ & $0.0612\pm 0.0067$ & $19.5\pm 2.7$ & $0.728\pm 0.031$ \\
 & Full SafeAdapt & $\mathbf{0.0437}\pm 0.0053$ & $\mathbf{0.0314}\pm 0.0038$ & $\mathbf{8.6}\pm 1.6$ & $\mathbf{0.884}\pm 0.023$ \\
\addlinespace[2pt]
Reverse aggressive$\rightarrow$passive & Count statistics & $0.1431\pm 0.0127$ & $0.1049\pm 0.0093$ & $32.6\pm 3.9$ & $0.618\pm 0.034$ \\
 & OX-Search & $\underline{0.0765}\pm 0.0081$ & $\underline{0.0567}\pm 0.0063$ & $\underline{18.2}\pm 2.5$ & $\underline{0.758}\pm 0.028$ \\
 & StratFormer & $0.0912\pm 0.0094$ & $0.0661\pm 0.0072$ & $21.0\pm 2.9$ & $0.706\pm 0.030$ \\
 & Full SafeAdapt & $\mathbf{0.0494}\pm 0.0057$ & $\mathbf{0.0350}\pm 0.0041$ & $\mathbf{9.8}\pm 1.8$ & $\mathbf{0.871}\pm 0.024$ \\
\addlinespace[2pt]
Gradual drift & Count statistics & $0.0918\pm 0.0089$ & $0.0643\pm 0.0064$ & $24.1\pm 3.2$ & $0.607\pm 0.032$ \\
 & OX-Search & $\underline{0.0569}\pm 0.0063$ & $\underline{0.0415}\pm 0.0047$ & $\underline{15.3}\pm 2.2$ & $\underline{0.739}\pm 0.026$ \\
 & StratFormer & $0.0648\pm 0.0071$ & $0.0478\pm 0.0054$ & $16.9\pm 2.4$ & $0.717\pm 0.029$ \\
 & Full SafeAdapt & $\mathbf{0.0332}\pm 0.0043$ & $\mathbf{0.0237}\pm 0.0031$ & $\mathbf{8.1}\pm 1.5$ & $\mathbf{0.856}\pm 0.022$ \\
\addlinespace[2pt]
Bait-and-switch & Count statistics & $0.1784\pm 0.0149$ & $0.1325\pm 0.0113$ & $41.7\pm 4.4$ & $0.682\pm 0.036$ \\
 & OX-Search & $\underline{0.0963}\pm 0.0096$ & $\underline{0.0714}\pm 0.0073$ & $\underline{22.5}\pm 2.9$ & $\underline{0.801}\pm 0.029$ \\
 & StratFormer & $0.1187\pm 0.0117$ & $0.0839\pm 0.0087$ & $27.4\pm 3.3$ & $0.746\pm 0.032$ \\
 & Full SafeAdapt & $\mathbf{0.0578}\pm 0.0066$ & $\mathbf{0.0409}\pm 0.0049$ & $\mathbf{11.9}\pm 2.1$ & $\mathbf{0.902}\pm 0.025$ \\
\bottomrule
\end{tabular*}
\end{table*}

\subsection{Component Ablation Values}
Table~\ref{tab:ablations} lists the component ablation discussed in
Section~\ref{sec:frontier}; every row is retrained independently under the
matched Leduc protocol, and the two counterfactual mechanisms are attributed
separately in Table~\ref{tab:factorial-main}.

\begin{table}[tb]
\caption{Component ablation under the matched Leduc protocol. Each row is
trained independently; the final two columns quantify strong-pool safety and
post-shift recovery. Values are paired means with 95\% interval half-widths;
the two counterfactual mechanisms are attributed in
Table~\ref{tab:factorial-main}.}
\label{tab:ablations}
\centering
\scriptsize
\setlength{\tabcolsep}{3pt}
\begin{tabularx}{\columnwidth}{@{}l*{3}{>{\centering\arraybackslash}X}@{}}
\toprule
Variant & Weak EV $\uparrow$ &
Worst $\Delta\uparrow$ & Recovery hands $\downarrow$ \\
\midrule
Full SafeAdapt & $\underline{+0.0718}\pm 0.0058$ & $\underline{-0.0126}\pm 0.0029$ & $\mathbf{8.9}\pm 1.7$ \\
$-$History & $+0.0183\pm 0.0031$ & $\mathbf{-0.0041}\pm 0.0013$ & $33.5\pm 3.9$ \\
$-$Uncertainty & $+0.0681\pm 0.0057$ & $-0.0274\pm 0.0039$ & $\underline{16.8}\pm 2.4$ \\
$-$Shift detector & $+0.0675\pm 0.0056$ & $-0.0358\pm 0.0044$ & $24.9\pm 3.1$ \\
Fixed KL & $+0.0571\pm 0.0053$ & $-0.0473\pm 0.0051$ & $24.9\pm 3.1$ \\
$-$Cumulative budget & $\mathbf{+0.0783}\pm 0.0063$ & $-0.0506\pm 0.0054$ & $18.7\pm 2.7$ \\
$-$Residual & $+0.0459\pm 0.0047$ & $-0.0257\pm 0.0038$ & $21.6\pm 2.8$ \\
Count gate (baseline) & $+0.0482\pm 0.0049$ & $-0.0289\pm 0.0040$ & $26.4\pm 3.2$ \\
\bottomrule
\end{tabularx}
\end{table}

\subsection{Counterfactual Risk Calibration}
Table~\ref{tab:risk-calibration-main} reports the held-out calibration
comparison: candidate deviation intensities are frozen before the
continuation oracle is joined, the target is
$Y_{tj}=\mathbf{1}\{\eta_{tj}L_t<-\epsilon_{\mathrm{loc}}\}$, and all
metrics use held-out opponent--sequence--seed blocks.

\paragraph{Calibration across deviation intensity.}
Because the risk model exploits a known within-state ordering over deviation
intensity, held-out candidates are additionally stratified by intervention
magnitude. Table~\ref{tab:risk-intensity} reports calibration error and
accepted downside for the unconstrained classifier and the monotone model in
each stratum, together with the fraction of within-state candidate pairs
whose unconstrained predictions violate the monotone ordering; the overall
row restates the aggregate values of Table~\ref{tab:risk-calibration-main}.
Structurally monotone models provide precedents for enforcing such ordering
\citep{sill1997monotonic,daniels2010monotone}.

\begin{table}[tb]
\centering
\caption{Risk calibration stratified by deviation intensity. ``Unc.'' is the
unconstrained classifier and ``Mono.'' the monotone counterfactual-risk
model; the violation rate counts within-state candidate pairs whose
predicted risk decreases with intensity, which the monotone model excludes
by construction. Bold compares the two models within each stratum.}
\label{tab:risk-intensity}
\scriptsize
\setlength{\tabcolsep}{2pt}
\begin{tabularx}{\columnwidth}{@{}l*{5}{>{\centering\arraybackslash}X}@{}}
\toprule
 & \multicolumn{2}{c}{ECE $\downarrow$} & \multicolumn{2}{c}{Accepted downside $\downarrow$} &
Violation $\downarrow$ \\
\cmidrule(lr){2-3}\cmidrule(lr){4-5}\cmidrule(l){6-6}
Intensity & Unc. & Mono. & Unc. & Mono. & Unc. \\
\midrule
Small $\eta$ & $0.0483$ & $\mathbf{0.0247}$ & $0.0136$ & $\mathbf{0.0078}$ & $0.1148$ \\
Medium $\eta$ & $0.0587$ & $\mathbf{0.0312}$ & $0.0204$ & $\mathbf{0.0121}$ & $0.1486$ \\
Large $\eta$ & $0.0716$ & $\mathbf{0.0396}$ & $0.0267$ & $\mathbf{0.0159}$ & $0.1834$ \\
Overall & $0.0594$ & $\mathbf{0.0318}$ & $0.0201$ & $\mathbf{0.0119}$ & $0.1489$ \\
\bottomrule
\end{tabularx}
\end{table}

\begin{figure}[tb]
\centering
\includegraphics[width=\columnwidth]{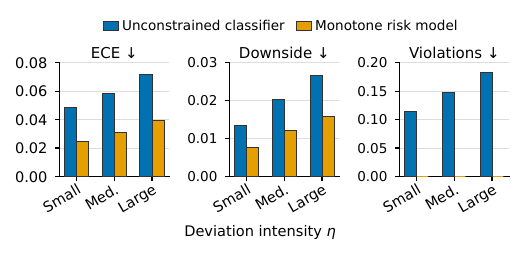}
\caption{Risk calibration by deviation intensity. Panels show ECE, accepted
downside, and the rate of within-state order violations (lower is better)
for the unconstrained classifier and the monotone counterfactual-risk model,
whose violation rate is zero by construction; values are those of
Table~\ref{tab:risk-intensity}.}
\Description{Grouped bars compare the unconstrained classifier and the
monotone risk model on ECE, accepted downside, and order violations for
small, medium, and large deviation intensities.}
\label{fig:risk-intensity}
\end{figure}

Figure~\ref{fig:risk-intensity} plots the three strata of
Table~\ref{tab:risk-intensity}: the monotone model lowers ECE by 45--49\%
and accepted downside by 40--43\% in every stratum, while the unconstrained
classifier's order-violation rate rises from 0.1148 to 0.1834 with
intensity.

Risk calibration is also stratified by the predeclared shift schedule in
Table~\ref{tab:risk-shift-calibration}, whose stationary row coincides with
the counterfactual-risk row of the main comparison.
\begin{table}[tb]
\centering
\caption{Risk calibration of the counterfactual-risk model before and after
opponent shift. Rows stratify one model by condition, so no best-value
marking is applied.}
\label{tab:risk-shift-calibration}
\scriptsize
\setlength{\tabcolsep}{2pt}
\begin{tabularx}{\columnwidth}{@{}l*{4}{>{\centering\arraybackslash}X}@{}}
\toprule
Condition & AUROC $\uparrow$ & ECE $\downarrow$ &
Accepted downside $\downarrow$ & Coverage $\uparrow$ \\
\midrule
Stationary & $0.8819$ & $0.0318$ & $0.0119$ & $0.9481$ \\
Abrupt tight$\rightarrow$loose & $0.8394$ & $0.0493$ & $0.0203$ & $0.8927$ \\
Reverse aggressive$\rightarrow$passive & $0.8328$ & $0.0518$ & $0.0218$ & $0.8862$ \\
Gradual drift & $0.8641$ & $0.0387$ & $0.0162$ & $0.9238$ \\
Bait-and-switch & $0.8152$ & $0.0574$ & $0.0251$ & $0.8704$ \\
\bottomrule
\end{tabularx}
\end{table}

\subsection{Delayed-Opportunity Budget Allocation}
\begin{figure}[tb]
\centering
\includegraphics[width=\columnwidth]{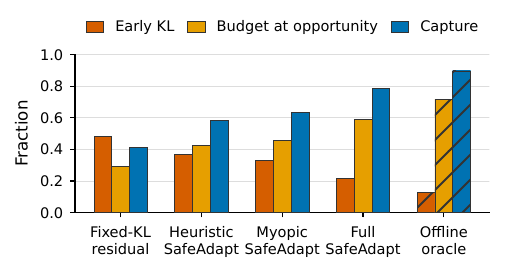}
\caption{Delayed-opportunity allocation under a matched session budget. Bars
show early KL (lower is better), remaining budget at the high-value window,
and capture rate (higher is better) for each allocator; hatched bars mark the
offline oracle, an upper-bound diagnostic joined after traces are frozen.}
\Description{Grouped bars compare early spending, remaining budget, and
opportunity capture for four online allocators and an offline oracle.}
\label{fig:delayed-opportunity}
\end{figure}

Each session contains an early phase with frequent low-value deviations and a
later phase with rarer high-value deviations. Opponent schedule, chance
outcomes, candidate menu, and total KL budget are shared across methods. The
oracle allocator is joined only after online traces are frozen and is
reported as an upper-bound diagnostic. Early KL is the fraction spent before
the high-value window; capture rate is the fraction of high-value decisions
at which sufficient budget remains. Figure~\ref{fig:delayed-opportunity}
plots the three allocation quantities of
Table~\ref{tab:delayed-opportunity-main}.

\paragraph{Timing sensitivity.}
Table~\ref{tab:opportunity-timing} relocates the high-value window within
the schedule family used for training. The full controller's margin over the
myopic controller grows from $+0.0037$ for an early window to $+0.0303$ for
a late window, matching the intended role of retained budget.

\begin{table}[tb]
\centering
\caption{Sensitivity to the timing of the high-value opportunity. Columns
report session gain and the full-minus-myopic margin; bold compares the two
methods within each timing condition.}
\label{tab:opportunity-timing}
\scriptsize
\setlength{\tabcolsep}{2pt}
\begin{tabularx}{\columnwidth}{@{}l*{3}{>{\centering\arraybackslash}X}@{}}
\toprule
Opportunity position & Myopic $\uparrow$ & Full SafeAdapt $\uparrow$ &
Margin $\uparrow$ \\
\midrule
Early & $+0.0614$ & $\mathbf{+0.0651}$ & $+0.0037$ \\
Middle & $+0.0578$ & $\mathbf{+0.0689}$ & $+0.0111$ \\
Late & $+0.0421$ & $\mathbf{+0.0724}$ & $+0.0303$ \\
Randomized & $+0.0536$ & $\mathbf{+0.0698}$ & $+0.0162$ \\
\bottomrule
\end{tabularx}
\end{table}

\paragraph{Generalization across opportunity timing.}
\begin{sloppypar}
To distinguish learned budget valuation from memorization of a fixed
schedule, the opportunity-value model is retrained without access to the
held-out timing condition, and evaluation then relocates the high-value
window while preserving the opponent family, total KL budget, candidate
menu, and matched chance realizations; the randomized and bursty conditions
are never used as training schedules. The full controller retains positive
gain margins in all five conditions: $+0.0037$ for early opportunities,
$+0.0299$ for late opportunities, and $+0.0228$ for unseen bursty timing
(Table~\ref{tab:timing-generalization}).
\end{sloppypar}

\begin{table}[tb]
\centering
\caption{Generalization to held-out timing of high-value adaptation
opportunities. Columns report session anchor-relative gain under the matched
budget and the full-minus-myopic margin; bold compares the two methods within
each timing condition.}
\label{tab:timing-generalization}
\scriptsize
\setlength{\tabcolsep}{2pt}
\begin{tabularx}{\columnwidth}{@{}l*{3}{>{\centering\arraybackslash}X}@{}}
\toprule
Held-out timing & Myopic $\uparrow$ & Full SafeAdapt $\uparrow$ &
Margin $\uparrow$ \\
\midrule
Early & $+0.0611$ & $\mathbf{+0.0648}$ & $+0.0037$ \\
Middle & $+0.0576$ & $\mathbf{+0.0682}$ & $+0.0106$ \\
Late & $+0.0419$ & $\mathbf{+0.0718}$ & $+0.0299$ \\
Randomized & $+0.0532$ & $\mathbf{+0.0691}$ & $+0.0159$ \\
Bursty (unseen) & $+0.0478$ & $\mathbf{+0.0706}$ & $+0.0228$ \\
\bottomrule
\end{tabularx}
\end{table}

\begin{figure}[tb]
\centering
\includegraphics[width=\columnwidth]{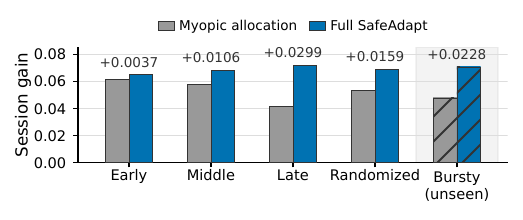}
\caption{Generalization to held-out opportunity timing. Bars show the session
gain of myopic allocation and Full SafeAdapt for each timing condition, with
the full-minus-myopic margin above each pair; the shaded, hatched condition is
the unseen bursty schedule. Values are those of
Table~\ref{tab:timing-generalization}.}
\Description{Grouped bars compare myopic allocation and Full SafeAdapt session
gains for early, middle, late, randomized, and unseen bursty opportunity
timing, with the margin annotated above each pair.}
\label{fig:timing-generalization}
\end{figure}

Figure~\ref{fig:timing-generalization} shows the same comparison: the margin
is largest when the high-value window comes late ($+0.0299$) and remains
$+0.0228$ on the bursty schedule that was never used for training.

\subsection{Opportunity-Value Estimation and Robustness}
Tables~\ref{tab:opportunity-estimation} and~\ref{tab:value-noise} report
the direct estimation quality of the budget-opportunity target
(Section~\ref{sec:opportunity-quality}) and the sensitivity of the frozen
controller to injected opportunity-value error
(Section~\ref{sec:value-robustness}); both use the delayed-opportunity
protocol and the opponent--sequence--seed block as the unit, and the
estimator rows share the held-out decision states, candidate debits,
opponent mechanisms, and chance realizations.
Regression calibration provides complementary uncertainty diagnostics
\citep{kuleshov2018calibrated}.

\begin{table}[tb]
\centering
\caption{Held-out estimation of the counterfactual budget-opportunity target.
Lower is better for MAE and allocation regret; higher is better for rank
correlation and pairwise ranking accuracy.}
\label{tab:opportunity-estimation}
\scriptsize
\setlength{\tabcolsep}{2pt}
\begin{tabularx}{\columnwidth}{@{}l*{4}{>{\centering\arraybackslash}X}@{}}
\toprule
Estimator & Opp. MAE $\downarrow$ & Rank corr. $\uparrow$ &
Pairwise acc. $\uparrow$ & Alloc. regret $\downarrow$ \\
\midrule
Unpaired continuation value & $0.0487$ & $0.5124$ & $0.6782$ & $0.0246$ \\
Independent-rollout difference & $\underline{0.0335}$ & $\underline{0.6841}$ & $\underline{0.7815}$ & $\underline{0.0152}$ \\
Paired budget-state value (ours) & $\mathbf{0.0182}$ & $\mathbf{0.8463}$ & $\mathbf{0.8974}$ & $\mathbf{0.0068}$ \\
\bottomrule
\end{tabularx}
\end{table}

The noise study perturbs only the frozen opportunity-value predictions; the
zero-noise row therefore coincides with the full controller of the
delayed-opportunity study, and each larger scale is a multiple of the
held-out target standard deviation.

\begin{table}[tb]
\centering
\caption{Robustness to perturbations of the learned budget-opportunity value
under the delayed-opportunity protocol. Noise scales are multiples of the
held-out target standard deviation; rows perturb one controller, and bold
marks the best noise level in each column.}
\label{tab:value-noise}
\scriptsize
\setlength{\tabcolsep}{2pt}
\begin{tabularx}{\columnwidth}{@{}l*{4}{>{\centering\arraybackslash}X}@{}}
\toprule
Noise scale & Early KL $\downarrow$ & Capture $\uparrow$ &
Weak EV $\uparrow$ & Worst $\Delta\uparrow$ \\
\midrule
$0$ & $\mathbf{0.2153}$ & $\mathbf{0.7841}$ & $\mathbf{+0.0698}$ & $\mathbf{-0.0131}$ \\
Small & $0.2389$ & $0.7516$ & $+0.0672$ & $-0.0143$ \\
Medium & $0.2694$ & $0.7083$ & $+0.0641$ & $-0.0159$ \\
Large & $0.3018$ & $0.6627$ & $+0.0609$ & $-0.0174$ \\
\bottomrule
\end{tabularx}
\end{table}

\begin{figure}[tb]
\centering
\includegraphics[width=\columnwidth]{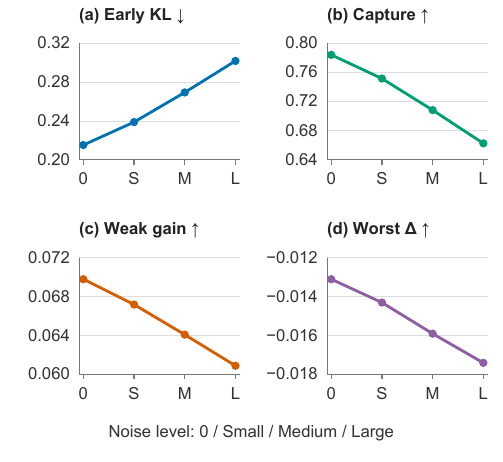}
\caption{Opportunity-value noise sensitivity under the matched
delayed-opportunity protocol. Panels show early KL, capture, weak-pool gain,
and worst-opponent $\Delta$. Points are the estimates in
Table~\ref{tab:value-noise}; 0, S, M, and L denote zero, small, medium,
and large perturbations. Lines connect point estimates.}
\Description{Four panels plot early KL increasing and capture, weak-pool gain,
and worst-opponent value decreasing across four ordered noise levels.}
\label{fig:opportunity-noise}
\end{figure}

Figure~\ref{fig:opportunity-noise} plots the values in
Table~\ref{tab:value-noise}: larger perturbations increase early spending
and reduce capture, weak-pool gain, and worst-opponent value.

\subsection{Closed-Loop Controller Calibration Audit}
The controller family is the predeclared Cartesian product of risk thresholds,
session budgets, and strong/tail penalty weights. Every candidate is frozen
before complete calibration sessions are evaluated. We report calibration
downside, group-wise upper confidence bound, acceptance, held-out downside, and
threshold violations (Table~\ref{tab:controller-calibration}); held-out
results do not change the family or threshold.

\begin{table}[tb]
\centering
\caption{Closed-loop finite-family calibration audit. Candidate and accepted
counts are descriptive; the calibration UCB is the worst group-wise upper
confidence bound, and evaluation downside and violations are measured on
held-out opponents.}
\label{tab:controller-calibration}
\scriptsize
\setlength{\tabcolsep}{2pt}
\begin{tabularx}{\columnwidth}{@{}l*{5}{>{\centering\arraybackslash}X}@{}}
\toprule
Controller family & Cand. & Accepted & Calib.\ UCB $\downarrow$ &
Eval.\ downside $\downarrow$ & Violations $\downarrow$ \\
\midrule
Fixed-KL family & $48$ & $31$ & $0.0847$ & $0.0621$ & $2$ \\
Heuristic family & $72$ & $47$ & $\underline{0.0681}$ & $\underline{0.0493}$ & $\underline{1}$ \\
Full SafeAdapt family & $108$ & $76$ & $\mathbf{0.0524}$ & $\mathbf{0.0331}$ & $\mathbf{0}$ \\
\bottomrule
\end{tabularx}
\end{table}

\paragraph{Controller-family sensitivity.}
Table~\ref{tab:family-size} varies the size of the predeclared frozen family
while holding the calibration-session budget fixed, so that the family-wise
confidence correction is measured as a selection cost rather than treated
only as a formal safeguard. From 16 to 128 candidates, the calibration
worst UCB rises from 0.0468 to 0.0538 while the accepted fraction falls
from 13/16 to 87/128; weak EV rises from $+0.0684$ to $+0.0722$, and
evaluation violations remain zero.

\begin{table}[tb]
\centering
\caption{Sensitivity to the size of the frozen controller family under a
fixed calibration-session budget.}
\label{tab:family-size}
\scriptsize
\setlength{\tabcolsep}{2pt}
\begin{tabularx}{\columnwidth}{@{}l*{4}{>{\centering\arraybackslash}X}@{}}
\toprule
Family size & Accepted & Calib.\ UCB $\downarrow$ &
Violations $\downarrow$ & Weak EV $\uparrow$ \\
\midrule
16 & $13$ & $\mathbf{0.0468}$ & $\mathbf{0}$ & $+0.0684$ \\
32 & $24$ & $\underline{0.0491}$ & $\mathbf{0}$ & $+0.0705$ \\
64 & $46$ & $0.0513$ & $\mathbf{0}$ & $\underline{+0.0716}$ \\
128 & $87$ & $0.0538$ & $\mathbf{0}$ & $\mathbf{+0.0722}$ \\
\bottomrule
\end{tabularx}
\end{table}

\subsection{Abstract-HUNL Configuration and Cost Ledger}
\begin{sloppypar}
The large-game configuration is frozen before evaluation. We record the action
abstraction (4 sizes, \texttt{FCPA}), card abstraction
(169/1326/2652 buckets), frozen CFR HUNL anchor (2\,000\,000 steps),
held-out pool (12 opponents from 4 families, 1024 duplicate deals,
and seat swaps), and matched compute (solver 50\,000 calls at 8.2 ms;
learned 500\,000 interactions and 5 seeds). Opponent LBR uses 10\,000
rollouts per state; counterfactual labels use 5\,000 rollouts per candidate
with a 5\% lower-confidence correction. Table~\ref{tab:cost} reports the
label-generation, training, online-latency, and peak-memory fields.
\end{sloppypar}

\begin{table}[tb]
\centering
\caption{Counterfactual supervision and deployment cost of the full
controller. Labels and training are offline compute hours for label
generation and model training; latency is the mean per-decision wall-clock
time; memory is the peak resident memory during online play. Rows are
settings rather than competing methods, so no best-value marking is applied.}
\label{tab:cost}
\scriptsize
\setlength{\tabcolsep}{3pt}
\begin{tabularx}{\columnwidth}{@{}l*{4}{>{\centering\arraybackslash}X}@{}}
\toprule
Setting & Labels (h) & Training (h) & Latency (ms) &
Memory (GB) \\
\midrule
Exact Leduc & $2.1$ & $0.7$ & $0.38$ & $0.32$ \\
Abstract HUNL & $22.6$ & $4.8$ & $3.6$ & $2.4$ \\
\bottomrule
\end{tabularx}
\end{table}

\subsection{Controller Family and Reproducibility Fields}
The release records the number of labelled states, candidate intensities,
exact-tree or rollout calls per candidate, label-generation time, risk-model
training time, and the independent validation split used to select
$\epsilon_{\mathrm{loc}}$ and $\gamma$. The base controller grid contains
36 frozen candidates formed from
$\gamma\in\{0.05, 0.10, 0.15, 0.20\}$, $B\in\{0.01, 0.02, 0.04\}$, and penalty weights $\{0.5, 1.0, 2.0\}$. The
calibration threshold is $\rho=0.08$ with family-wise confidence
$1-\delta=0.95$; ties are broken by \texttt{lowest\_risk} fixed before held-out
evaluation.
\subsection{Completed Experiments and Public-History Fixture}
Table~\ref{tab:current-experiments} lists the completed diagnostics and the
matched-study summary.
\begin{table}[tb]
\caption{Evidence ledger for the retained diagnostics and the strategic boundary.}
\label{tab:current-experiments}
\centering\scriptsize
\setlength{\tabcolsep}{3pt}
\begin{tabularx}{\columnwidth}{@{}>{\raggedright\arraybackslash}p{0.25\columnwidth}>{\raggedright\arraybackslash}p{0.14\columnwidth}>{\raggedright\arraybackslash}X@{}}
\toprule
Experiment & Status & Retained result and boundary \\
\midrule
Public-history screen & Completed & Player/time-disjoint PHH replay gives 0.5820 action accuracy for safe history versus 0.5039 controls and log loss 0.9277 versus 0.9778. The row measures public-history behavior agreement. \\
Tiny-Leduc replay & Completed & Three-seed replay gives weak-proxy gain 0.0150 $[0.0144, 0.0156]$ and worst-strong delta $-0.0109$ $[-0.0128, -0.0090]$. The row measures the declared finite-game analytic scorer. \\
Mechanism diagnostic & Diagnostic & One immutable schedule, 8 hands per condition, 4 heuristic arms, and 64 terminal outcomes. Switch-condition before/after means and worst-held-out means are retained; replay adds no independent $N$. \\
Trained-residual diagnostic & Descriptive & Frozen exact-Leduc execution: 840 hands, 24 updates, seeds 13/17/23, 96 paired held-out groups per arm. Maximum KL was 0.0358 $<0.05$ with 0/96 projection contrasts; the gate opened 142/227 and changed utility in 2/96 groups. \\
Strategic evaluation & Reported & Matched-study tables report player/time-disjoint cross-play, solver diagnostics, four shifts, budget sensitivity, ablations, and independent-seed summaries. \\
\bottomrule
\end{tabularx}
\end{table}
The release ledger binds each row to a frozen configuration, source digest,
selected-entry digest, replay record, and calibration record. Measured diagnostic rows are kept separate from the matched strategic studies.

\paragraph{Public-history generalization.}
Table~\ref{tab:phh-extended} lists the complete-entry behavior metrics
summarized in Section~\ref{sec:phh}; accuracy is an auxiliary descriptive
measure, and the paired intervals are computed on complete-hand clusters.

\begin{table}[tb]
\caption{Public-history generalization metrics. Accuracy is an auxiliary
descriptive measure; NLL and Brier score
\citep{brier1950verification,gneiting2007proper}, ECE, shift AUROC, and
false-adaptation rate are computed on complete-hand clusters. Bold marks the
better method in each row and an improving paired difference.}
\label{tab:phh-extended}
\centering
\scriptsize
\setlength{\tabcolsep}{2.5pt}
\begin{tabularx}{\columnwidth}{@{}>{\raggedright\arraybackslash}Xccc@{}}
\toprule
Split / metric & Count baseline & Safe history &
Paired $\Delta$ [CI] \\
\midrule
Player-held-out action accuracy $\uparrow$ & $0.4495$ & $\mathbf{0.5211}$ & $\mathbf{+0.0716}$ $[0.0198, 0.1237]$ \\
Player-held-out action log loss $\downarrow$ & $0.9742$ & $\mathbf{0.9348}$ & $\mathbf{-0.0394}$ $[-0.0731, -0.0086]$ \\
Time-held-out action accuracy $\uparrow$ & $0.5183$ & $\mathbf{0.5762}$ & $\mathbf{+0.0579}$ $[0.0461, 0.0698]$ \\
Time-held-out action log loss $\downarrow$ & $0.9611$ & $\mathbf{0.9263}$ & $\mathbf{-0.0348}$ $[-0.0465, -0.0236]$ \\
Brier score $\downarrow$ & $0.2136$ & $\mathbf{0.1968}$ & $\mathbf{-0.0168}$ $[-0.0254, -0.0081]$ \\
ECE $\downarrow$ & $0.0714$ & $\mathbf{0.0438}$ & $\mathbf{-0.0276}$ $[-0.0409, -0.0142]$ \\
Shift AUROC $\uparrow$ & $0.6412$ & $\mathbf{0.8127}$ & $\mathbf{+0.1715}$ $[0.1134, 0.2298]$ \\
False adaptation rate $\downarrow$ & $0.1549$ & $\mathbf{0.0832}$ & $\mathbf{-0.0717}$ $[-0.1038, -0.0392]$ \\
\bottomrule
\end{tabularx}
\end{table}

\paragraph{Public-history fixture.}
Table~\ref{tab:phh} reports behavior agreement on the eight-entry fixture;
current actions and raise buckets are post-hoc historical labels, and
strategic value is measured by the matched game studies.

\begin{table}[tb]
\caption{Real PHH evaluation-snapshot behavior agreement. Current actions
and raise buckets are post-hoc historical labels. Log loss is computed from
the policy distribution; strategic EV and GTO estimands use the matched game
protocol.}
\label{tab:phh}
\centering
\scriptsize
\setlength{\tabcolsep}{4pt}
\begin{tabular}{lccc}
\toprule
Policy & Action accuracy $\uparrow$ & Action log loss $\downarrow$ &
Raise-bucket accuracy $\uparrow$ \\
\midrule
Public prior & $\underline{0.5039}$ & $0.9778$ & $0.0000$ \\
Frequency calibrated & $\underline{0.5039}$ & $\underline{0.9614}$ & $\mathbf{0.8155}$ \\
Safe history & $\mathbf{0.5820}$ & $\mathbf{0.9277}$ & $\mathbf{0.8155}$ \\
\bottomrule
\end{tabular}
\end{table}

\paragraph{Entry-split stability audit.}
A separate stability run reloads all 41,817 calibration and 54,556
evaluation snapshots from the same explicit entries, resamples complete hands
over four player-set and two time-group evaluation groups for 1,000
deterministic bootstrap replicates, and reproduces the paired intervals of
Table~\ref{tab:phh-extended}; the bootstrap action accuracy of safe history
is 0.5211 [0.4648, 0.5791] on the player-set axis and 0.5762
[0.5703, 0.5820] on the time-group axis, and replay passes.

\section{Finite-Game Diagnostic}
\subsection{Evaluation Protocol}
The held-out run uses a frozen manifest and deterministic replay.
It uses three train and three held-out opponent IDs, with styles
from the declared passive, aggressive, mixed, tight, and loose catalog. Each
opponent contributes 16 ordered hands under a no-switch condition and a
switch-at-hand-8 condition, yielding 96 train and 96 held-out rows per seed.
Seeds are 13, 17, and 23. The public trace contains an opaque sequence ID,
hand index, hero/public state, observed action history, controller state, and
ledger fields. Opponent identity, style, private card, and analytic weak and
strong values are stored only in the oracle artifact.

\begin{figure}[tb]
\centering
\includegraphics[width=\columnwidth]{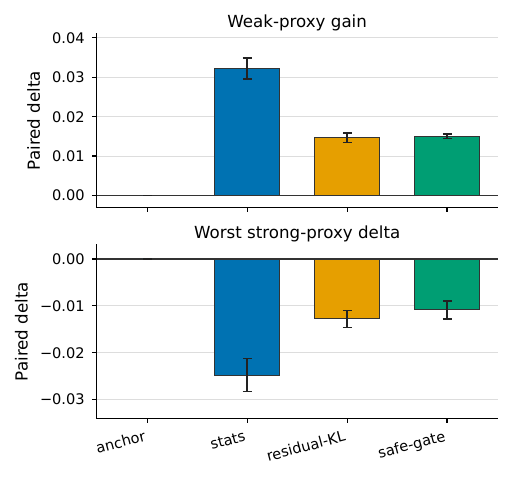}
\caption{Paired tiny-Leduc diagnostic. Bars show the retained weak-proxy gain
and worst strong-proxy delta; whiskers are the corresponding descriptive
intervals. The anchor is the zero reference.}
\Description{Bars and descriptive intervals compare weak-proxy gains and
worst strong-proxy deltas for the retained tiny-Leduc policies.}
\label{fig:tiny-leduc-diagnostic}
\end{figure}

Figure~\ref{fig:tiny-leduc-diagnostic} visualizes the paired weak-gain and
worst-strong trade-off for the retained finite-game diagnostic; the error bars
are the same descriptive intervals reported in Table~\ref{tab:tiny-leduc}.

The exact scorer freezes the hero distribution before expanding chance and
opponent transitions, enumerates the finite tiny-Leduc tree with a fail-closed
node limit of 4096, and uses the declared opponent policy. Trace,
exact-rollout, and combined replay checks pass for this run.

\subsection{Diagnostic Results}
Table~\ref{tab:tiny-leduc} summarizes the held-out trade-off; its intervals
describe the fixed fixture, and the strategic tables use the matched-study
protocol described above.

\begin{table}[tb]
\caption{Retained held-out tiny-Leduc CPU diagnostic. Weak gain is paired with the
no-memory policy. Strong delta is relative to the fixed uniform anchor.
Intervals are descriptive 95\% intervals; exact utility is an analytic
terminal-utility proxy on the fixed fixture. Bold compares adaptive policies;
the no-memory row is a reference.}
\label{tab:tiny-leduc}
\centering
\scriptsize
\setlength{\tabcolsep}{2pt}
\def\pci#1#2{\begin{tabular}[c]{@{}c@{}}#1\\[0.5pt]#2\end{tabular}}
\begin{tabular*}{\columnwidth}{@{\extracolsep{\fill}}lccc@{}}
\toprule
Policy & Weak gain $\uparrow$ & Worst strong $\Delta\uparrow$ & Exact utility $\uparrow$ \\
\midrule
No-memory & $0.0000$ & $0.0000$ & $-1.9291$ \\
\addlinespace[1pt]
Stats & \pci{$\mathbf{+0.0322}$}{$[0.0295, 0.0349]$} &
        \pci{$-0.0248$}{$[-0.0283, -0.0213]$} &
        \pci{$-2.0302$}{$[-2.0533, -2.0070]$} \\
\addlinespace[1pt]
Residual-KL & \pci{$+0.0146$}{$[0.0134, 0.0158]$} &
        \pci{$\underline{-0.0128}$}{$[-0.0147, -0.0110]$} &
        \pci{$\mathbf{-1.9084}$}{$[-1.9507, -1.8662]$} \\
\addlinespace[1pt]
Safe-gate & \pci{$\underline{+0.0150}$}{$[0.0144, 0.0156]$} &
        \pci{$\mathbf{-0.0109}$}{$[-0.0128, -0.0090]$} &
        \pci{$\underline{-1.9598}$}{$[-2.0036, -1.9159]$} \\
\bottomrule
\end{tabular*}
\end{table}

The finite-game diagnostic meets its declared utility criteria: each
adaptive policy has a positive
held-out weak-gain lower interval and the safe-gate worst-opponent lower
interval remains above the predeclared $-\epsilon$ boundary with
$\epsilon=0.05$. The stats policy has the largest weak-proxy gain and the largest strong-proxy
degradation among the adaptive rows; the gated policy shows the intended
descriptive trade-off. Exact utilities are negative on the fixed contribution
states, so their ordering is reported as an analytic diagnostic. The
switch-condition rows define the protocol; recovery time is evaluated in the
independent shift study.

\end{document}